\def\doublecolumn{1} %
\if 1\doublecolumn
  \documentclass[journal]{IEEEtran}
\else
  \documentclass[12pt, draftclsnofoot, onecolumn, letterpaper]{IEEEtran}
\fi

\def\blind{1} %

\usepackage[latin9]{inputenc}
\usepackage{algpseudocode}
\usepackage{algorithm}
\usepackage{array}

\usepackage{makecell} 
\usepackage{float}
\usepackage{mathtools}
\usepackage{amsmath}
\usepackage{amsthm}
\usepackage{amssymb}
\usepackage{graphicx}
\usepackage{wasysym}
\usepackage{setspace}
\usepackage{color}
\usepackage{bm}
\usepackage{cases}
\usepackage{tikz}
\usetikzlibrary{calc}
\usepackage{flowchart}
\usepackage{bbm}
\usepackage{adjustbox}
\usepackage{hhline}
\usepackage{makecell}

\usepackage{epsfig}
\usepackage{cite}
\usepackage{graphicx}
\usepackage{lipsum}
\usepackage{multirow}
\usepackage{array}
\usepackage{enumerate}
\usepackage{enumitem}
\usepackage{graphicx}
\usepackage{times}
\usepackage[framemethod=tikz]{mdframed}
\definecolor{cccolor}{rgb}{1,1,1}

\usepackage{subfig}
\usepackage{caption}
\newtheorem{remark}{Remark}

\usepackage{multicol,multirow,adjustbox}
\usepackage{booktabs}

\newcommand{\ie}[0]{\textit{i.e.}}

\usepackage[capitalize]{cleveref}
\crefname{section}{Sec.}{Secs.}
\Crefname{section}{Section}{Sections}
\crefname{table}{Tab.}{Tabs.}
\Crefname{table}{Table}{Tables}

\makeatletter
\def\@IEEEsectpunct{.\ \,}
\def\paragraph{\@startsection{paragraph}{4}{\z@}{1.2ex plus 1.1ex minus 0.5ex}%
{0ex}{\normalfont\normalsize\bfseries}}
\makeatother

\usepackage{pifont,xspace}
\newcommand{\cmark}{\ding{51}\xspace}%
\newcommand{\xmarkg}{\textcolor{lightgray}{\ding{55}}\xspace}%

\begin{document}

\title{
    Unveiling Hidden Visual Information: \\ A Reconstruction Attack Against Adversarial Visual Information Hiding
}

\if 1\blind
\author{Jonggyu Jang,~\IEEEmembership{Member,~IEEE}, Hyeonsu Lyu,~\IEEEmembership{Student Member,~IEEE}, Seongjin Hwang, and  Hyun~Jong~Yang,~\IEEEmembership{Member,~IEEE}
    \thanks{
    J. Jang, H. Lyu, S. Hwang, and H. J. Yang (corresponding author) are with Department of Electrical Engineering, Pohang University of Science and Technology (POSTECH), Pohang 37673, Republic of Korea, (e-mail: \{jgjang,hslyu4,sjh1753,hyunyang\}@postech.ac.kr). 
    J. Jang and H. Lyu are equally contributed. 
    }
}
\else
\author{Anonymous Submission
    }
\fi
\maketitle

\begin{abstract}
    This paper investigates the security vulnerabilities of adversarial-example-based image encryption by executing data reconstruction (DR) attacks on encrypted images.
    A representative image encryption method is the adversarial visual information hiding (AVIH), which uses type-I adversarial example training to protect gallery datasets used in image recognition tasks. In the AVIH method, the type-I adversarial example approach creates images that appear completely different but are still recognized by machines as the original ones. Additionally, the AVIH method can restore encrypted images to their original forms using a predefined private key generative model.
    For the best security, assigning a unique key to each image is recommended; however, storage limitations may necessitate some images sharing the same key model. This raises a crucial security question for AVIH: \textit{How many images can safely share the same key model without being compromised by a DR attack?}
    To address this question, we introduce a dual-strategy DR attack against the AVIH encryption method by incorporating \textit{(1) generative-adversarial loss} and \textit{(2) augmented identity loss}, which prevent DR from overfitting---an issue akin to that in machine learning.
    Our numerical results validate this approach through image recognition and re-identification benchmarks, demonstrating that our strategy can significantly enhance the quality of reconstructed images, thereby requiring fewer key-sharing encrypted images. Our source code to reproduce our results will be available soon.
\end{abstract}

\begin{figure}[t]
    \centering
    \includegraphics[width=\if 1\doublecolumn .85 \else 0.5 \fi\linewidth]{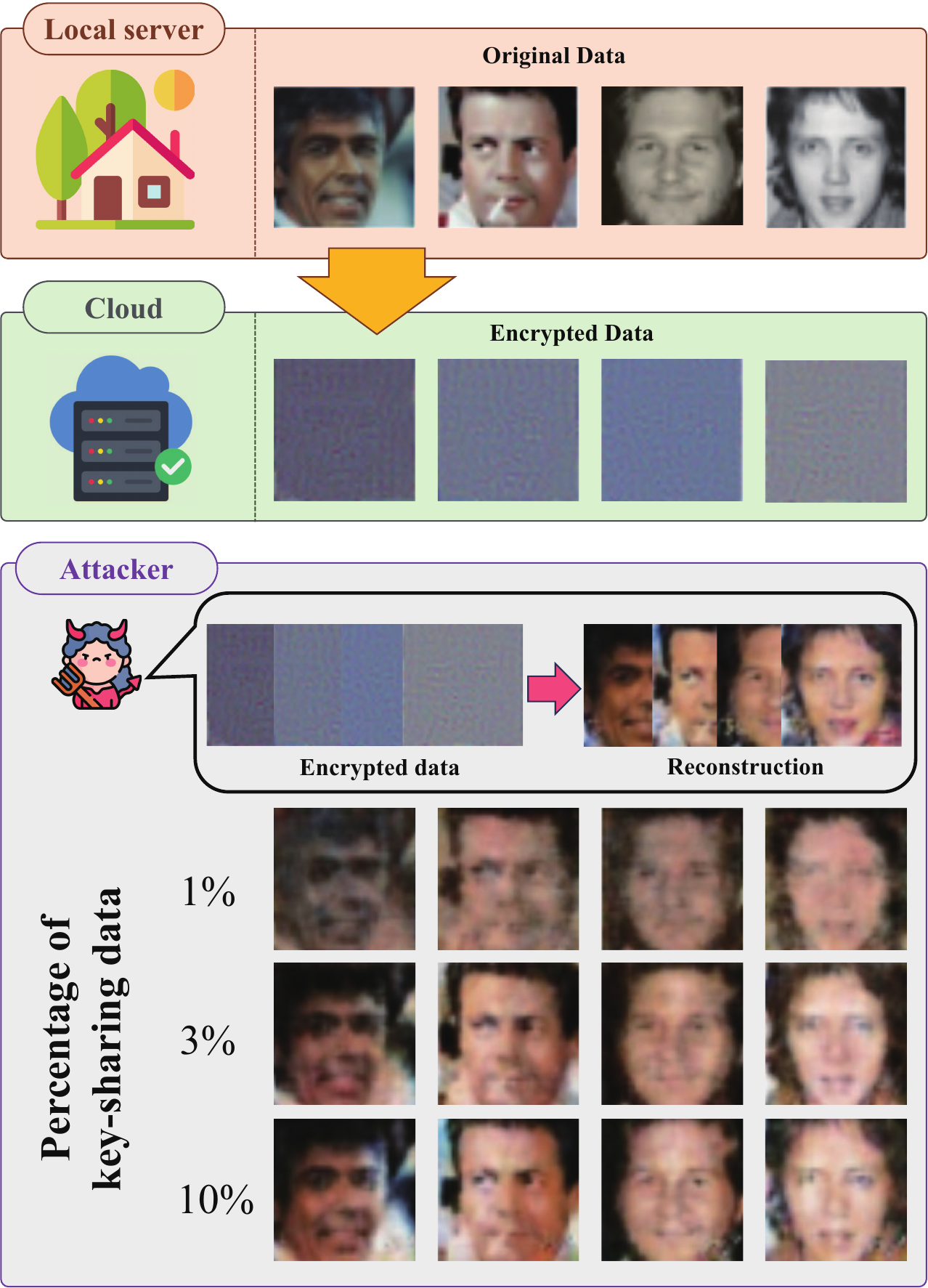}
    \caption{
        Examples of the encryption method and the proposed attacker model. 
        The local server stores the original image data. 
        In a cloud service system, the local server offloads the computation of image recognition tasks (such as face recognition in our example) to the cloud server. 
        Before sending raw data of the private images to the cloud server, the local server encrypts the original image data into a noisy image. 
        The service model then processes both the original and encrypted images similarly. 
        Our focus in this scenario is to highlight the privacy vulnerabilities of the encryption method through data reconstruction attacks.
    }
    \label{fig:introduction_motivation}
\end{figure}

\section{Introduction\label{sec:intro}}

\IEEEPARstart{M}{achine} learning has evolved from a groundbreaking innovation to a widely adopted and promising technology across numerous fields. One key characteristic of machine learning is its dependency on data; machine learning models are trained on data and often require additional user data for their application services. Recently, this dependency on data has raised significant privacy concerns~\cite{su2023hiding, chollet2024privacypreservingpersonalassistant}. In the most straightforward case, storing and processing facial or body images in public cloud services for machine learning applications can expose these images to unauthorized access and misuse~\cite{singh2017cloud}.

In such an inference scenario, a simple solution is to abstain from transmitting the data and conduct on-device computing~\cite{chollet2024privacypreservingpersonalassistant}. However, due to the limited computing power and battery capacity of mobile devices, it is not feasible to fully compute services locally. 
Henceforth, there is an increasing need for research into privacy-preserving machine learning techniques to address practical scenarios.

\begin{figure*}[!ht]
    \centering
    \includegraphics[width=0.95\linewidth]{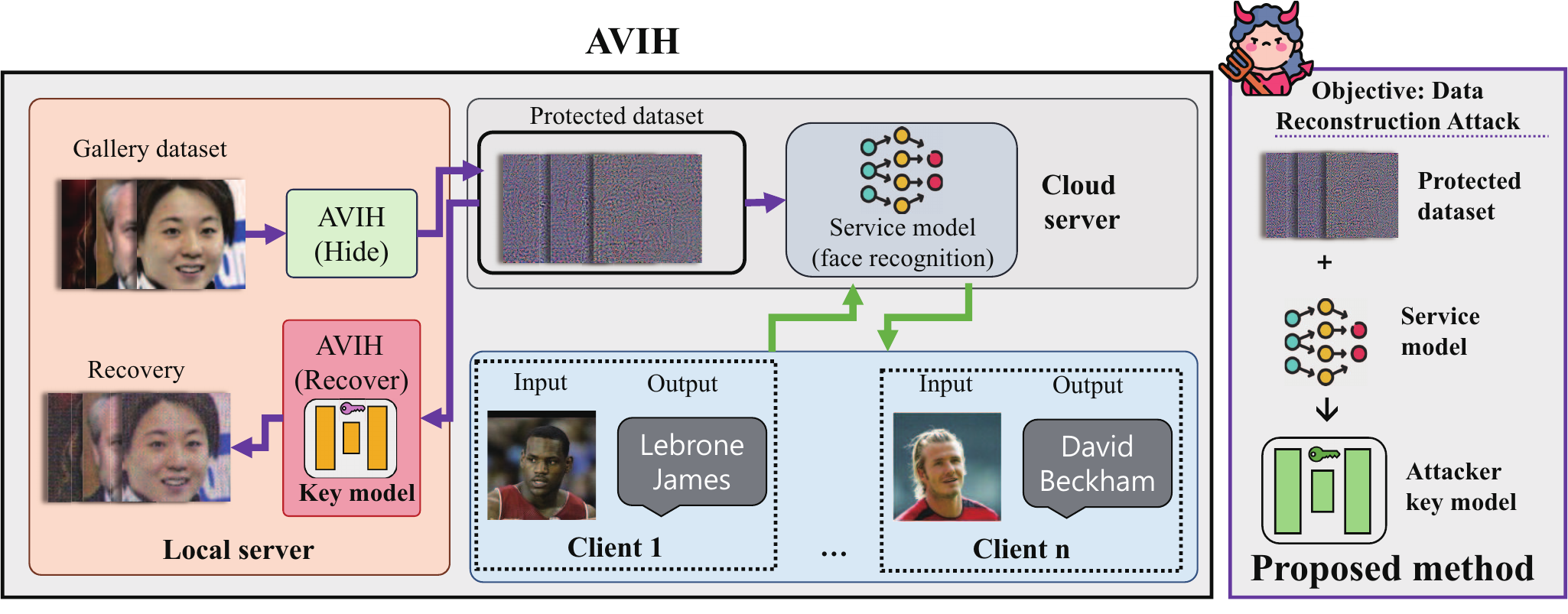}
    \caption{
    An illustration of the AVIH method and the objective of our work is shown. The left part demonstrates that the gallery set is safeguarded and provided to the DNN in the cloud server. The protected image shows altered visual information that is completely different from the original, yet remains accurately identifiable. In the local server, the protected images can be recovered using its key DNN model.
    The right part of the illustration highlights our proposed method, which aims to design \textbf{a replica of the key model} without accessing the actual key model.
    }
    \label{fig:AVIH_example_intro}
\end{figure*}

\paragraph*{Background}
This work focuses on security and privacy issues in cloud-based image recognition systems~\cite{he2020fastreid,he2021transreid}, as depicted in \cref{fig:introduction_motivation}.
In the cloud-based system, the gallery dataset should be accessible at the cloud server because image recognition tasks involve comparing gallery images with target images, where gallery images contain sensitive information and are often unencrypted.
Such leakage of the gallery dataset has a potential privacy vulnerability, where it is susceptible to simple cyber threats, as malicious attackers can directly access the sensitive face images of all users~\cite{su2023hiding}.
Hence, the local server aims to conceal the visual information before sending gallery images to cloud servers.

\paragraph*{Defensive methods}
Several defensive methods have been proposed to counteract the security and privacy risks associated with gallery datasets, including: 1) hiding visual information in noisy images~\cite{su2023hiding, zhu2018hidden}, 2) perceptual encryption~\cite{xiang2019visual, guo2019peid}, and 3) homomorphic encryption~\cite{lee2022low, kim2023optimized}\footnote{More comprehensive literature reviews are available in \cref{sec:related_works}}. Homomorphic encryption can guarantee the utility of DNN computation with strong security; however, it requires excessive computation time, making it impractical for cloud-based systems. Additionally, perceptual encryption necessitates re-training the service DNN, which significantly degrades inference quality after encryption.
Conversely, the AVIH (Adversarial Visual Information Hiding) encryption method~\cite{su2023hiding} encrypts gallery images into noisy images while preserving the output of the service model, as depicted in \cref{fig:introduction_motivation}. Therefore, the AVIH method maintains the performance of the service without any speed degradation. For these reasons, our focus is on the AVIH encryption method~\cite{su2023hiding}.
However, despite its significance, there has been no work analyzing the effectiveness of AVIH in securing information.

\paragraph*{Research question}
In the AVIH method, assigning a unique key to each image is recommended for optimal security; however, storage limitations may necessitate some images sharing the same key model.
Throughout this paper, we address the following research question regarding the practical use of the AVIH method for cloud-based inference systems:
\begin{mdframed}[outerlinecolor=black,outerlinewidth=1pt,linecolor=cccolor,middlelinewidth=1pt,roundcorner=0pt]
  \begin{center}
    \textbf{\textit{RQ: How securely can adversarial visual information hiding truly conceal visual information?}}
  \end{center}
\end{mdframed}
In the remainder of this paper, we aim to solve the above research question by executing a \textbf{data reconstruction} attack against the AVIH method.
Examples of the data reconstruction attack are depicted at the bottom of \cref{fig:introduction_motivation}.

\paragraph*{Data reconstruction attacks}
In previous studies~\cite{Nguyen_2023_CVPR, Tian_23_roleof, Zhou_23_boosting_MI_adv, hu2022membership, Fredrikson14, Fredrikson15, Hidano17, jang2023patchmi}, data reconstruction attacks have been proposed for recovering training data from trained neural networks. This concept arises from the widespread belief that trained DNNs can retain information about their training data, with a simple example being inferring the membership of data~\cite{hu2022membership}.
Beyond merely inferring membership, researchers have developed data reconstruction attack methods for simple classifiers, though these reconstructed images often lack photo-realism~\cite{Fredrikson14, Fredrikson15, Hidano17}. With the advent of generative models, new techniques have emerged to produce photo-realistic images using methods such as advanced identity loss leveraging logits~\cite{Nguyen_2023_CVPR}, supervised inversion~\cite{Tian_23_roleof}, and adversarial examples~\cite{Zhou_23_boosting_MI_adv}, where those methods focus on finding appropriate latent vectors.
Our study differentiates itself by not relying on pre-trained DNNs for finding latent vectors, leading to imperfect reconstructed images. 
Instead, we train an attacker key model that mimics the original key model, aiming to \textbf{exactly} reconstruct the original images.

\begin{figure}
    \centering
    \includegraphics[width=0.9\linewidth]{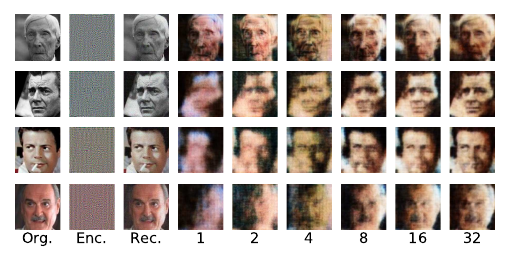}
    \caption{Original images, encrypted images, reconstructed images, and our attack results. The numbers below the images refer the number of images sharing the same key model.}
    \label{fig:example_reconstruction}
\end{figure}

\subsection{Our Findings}

In this paper, we aim to show that leveraging adversarial examples for visual information hiding~\cite{su2023hiding} is \textbf{unsafe}, as depicted in the right part of \cref{fig:AVIH_example_intro}.
To this end, we propose a data reconstruction attack on the encrypted gallery set without access to the \textbf{original key model}.
In the proposed method, we first randomly initialize an attacker key model. 
Then, we train the attacker key model to recover an image that consistently matches the service DNN output while ensuring photo-realism.
As a simple method, one can train the attacker key model with the identity loss between the service model outputs of encrypted data and the attacker key model's output. 
However, with only standard identity loss, the attacker key model's outputs are not similar to the original images. 
This is our main challenge in the generalization of the attacker key model, where overfitting (a concept similar to ordinary machine learning) occurs if few data share the same key model.

Our salient contributions are summarized as follows:
\begin{itemize}
    \item \textbf{Exact reconstruction attack. } To the best of the authors' knowledge, our work is the first to demonstrate that an exact data reconstruction attack works in practical scenarios.\footnote{As shown in previous studies~\cite{haim2022reconstructing}, given a DNN's output features or logits, one can exactly reverse-engineer the DNN to reconstruct the original training dataset, though a strong assumption (homogeneous neural network) is required.}
    \item \textbf{Resolving overfitting 1 (augmented identity loss). } To alleviate the overfitting issue in data reconstruction attacks, we propose augmented identity loss, which helps generalize the trained attacker key model.
    \item \textbf{Resolving overfitting 2 (generative-adversarial loss). } In addition to augmented identity loss, we propose a GAN-based data reconstruction attack to improve the generalization and photo-realism of the attacker outputs. Unlike previous studies~\cite{Nguyen_2023_CVPR, Tian_23_roleof, Zhou_23_boosting_MI_adv} that use pre-trained DNNs to find an appropriate latent vector, we train an auxiliary key model that mimics the original key model while constraining local patch-level similarity with an auxiliary dataset.
    \item \textbf{Vulnerabilities of the AVIH method. } We validate that images encrypted by the AVIH method can be reconstructed by the proposed method for various tasks such as face recognition, human re-identification, and vehicle identification. For example, \cref{fig:example_reconstruction} shows the results of the proposed method by changing the number of key-sharing images for a face recognition model. Furthermore, our ablation studies show that the proposed method can enhance the quality of data reconstruction, highlighting the necessity of stronger privacy-preserving methods.
\end{itemize}

\subsection{Organization}

The remaining parts of this paper are organized as follows.
In \cref{sec:related_works}, we provide comprehensive reviews of existing visual information hiding methods and corresponding security and privacy attacks.
Section \ref{sec:methodology} details the proposed approach for data reconstruction against the AVIH method, including a detailed review of the AVIH method.
Next, in \cref{sec:experiments} and Appendix \ref{sec:additional_experiments}, we present our experimental setup and results for face recognition scenarios and re-identification scenarios, respectively.
Finally, \cref{sec:discussion} concludes the paper with a discussion on the conclusion, limitations, extensibility, and future research directions.

\begin{figure*}
    \centering
    \includegraphics[width=\if 1\doublecolumn 0.7 \else 0.8 \fi \linewidth]{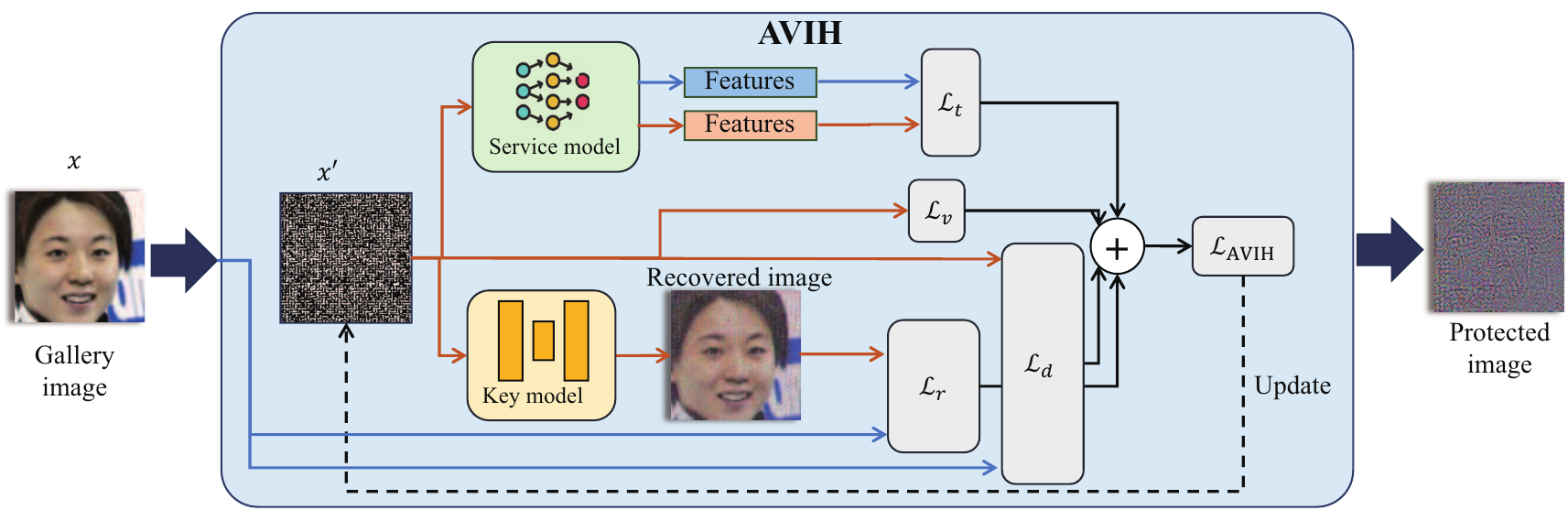}
    \caption{
    Adversarial visual information hiding method proposed in \cite{su2023hiding}. Given a key model, the AVIH method aims to generate type-I adversarial examples (noisy images) that produce very similar outputs for the target service model. To achieve this, the adversarial image $x'$ is modified to minimize task loss ($L_t$), difference loss ($L_d$), recovery loss ($L_r$), and variance consistency loss ($L_v$). The details of these losses will be presented in \cref{subsec:background}.
    }
    \label{fig:avih}
\end{figure*}

\section{Related Works on Hiding Visual Information}
\label{sec:related_works}

Several studies have focused on hiding visual information in machine learning tasks, particularly during the inference stage.

\paragraph*{Homomorphic encryption} 
Homomorphic encryption (HE) is a cryptographic technique that allows computations to be performed on encrypted data, maintaining privacy while still producing an encrypted result that, when decrypted, matches the result of operations performed on the original data. In~\cite{gilad2016cryptonets}, HE for deep neural networks was proposed. Extending HE to deeper neural networks, a low-complexity encryption method for DNNs was introduced in~\cite{lee2022privacy, lee2022low}. Although HE effectively prevents privacy leakage, the state-of-the-art HE remains extremely slow for computing large neural networks.

\paragraph*{Perceptual encryption}

In the inference stage, perceptual encryption (PE) has emerged as a promising method for finding a suitable encrypted domain for visual images. 
In~\cite{ding2020deepedn}, a cycle-GAN-based PE method was proposed for medical images, requiring encryption/decryption keys. 
In~\cite{sirichotedumrong2021gan}, a more advanced method eliminated the need for these keys, using the cycle-GAN model itself as a unique encryption key. However, the purported security of PE is questionable. Numerous studies have demonstrated that PE is highly vulnerable to data reconstruction attacks, which can effectively restore original images even against pixel-based encryption~\cite{chuman2023jigsaw} and learnable encryption~\cite{maungmaung2023generative}. This vulnerability underscores a critical flaw in PE methods, challenging their viability for robust security in practical applications.

\paragraph*{Adversarial examples} 
Adversarial examples are widely used in privacy-related deep learning technologies due to their versatility, such as for inserting watermarks in foundation models~\cite{tang2023watermarking, qiao2023novel}. For hiding visual information, unlike HE and PE, steganography~\cite{baluja2017hiding} and AVIH~\cite{su2023hiding} are practical approaches, as they can hide information or recover the original data with low computational complexity. 
More specifically, AVIH can guarantee the correctness of computational results. The variance-consistency loss used in AVIH can efficiently encrypt visual information while preserving the correctness of DNN computations without requiring any re-training process, which could lead to additional privacy leakage. 
However, privacy-preserving aspects of AVIH have been discussed heuristically and do not meet stringent privacy constraints. 
In this work, we present the first approach that executes data reconstruction attacks on the AVIH method.

\section{Methodology}
\label{sec:methodology}
In this work, our primary focus is to demonstrate that sharing the same key models for hiding visual information poses potential security and privacy vulnerabilities.
In this section, we first briefly review the AVIH method in \cref{subsec:background} and then propose our method in \cref{subsec:proposed}.

\subsection{Preliminary: AVIH Encryption}\label{subsec:background}

Consider a remote computing scenario where clients provide a gallery dataset and a target service model to a cloud server. If the original gallery dataset is sent to the server without encryption, the sensitive and private visual images that clients wish to keep confidential can be accessed by the server administrator. 
To address this, the AVIH method is proposed to secure the gallery dataset images while maintaining the service quality of the target service model.
To ensure the reconstruction of the original data from the encrypted data, clients have their private and secure key model, accessible only to them. For image recognition tasks, such as face recognition~\cite{deng2019arcface,kim2022adaface} or human re-identification~\cite{he2021transreid}, the encrypted gallery datasets can be used without a loss of accuracy. Additionally, while humans cannot recognize the original image from the encrypted version, the secure key model can accurately reconstruct the original image.

\paragraph*{Overall loss function}

In the AVIH method, the primary goal is to generate images that 1) preserve the service model's output, 2) destroy the image's structural information, and 3) guarantee recovery of the original image with a key model. 
From the original image $x_i$, we generate the image $x_i'$, where $i$ denotes the index of the gallery images. The generated image $x_i'$ is designed to minimize the following loss function:
\if 1\doublecolumn
\begin{multline}
\label{eq:loss_AVIH}
    L_{\text{AVIH}}(x_i,x_i') = L_{t}(x_i,x_i')  \\ 
    - \lambda_1 L_d(x_i, x_i') +  \lambda_2 L_v(x_i') + \lambda_3 L_r(x_i,x_i').
\end{multline}
\else
\begin{equation}
\label{eq:loss_AVIH}
    L_{\text{AVIH}}(x_i,x_i') = L_{t}(x_i,x_i')
    - \lambda_1 L_d(x_i, x_i') +  \lambda_2 L_v(x_i') + \lambda_3 L_r(x_i,x_i').
\end{equation}
\fi 

The details of the AVIH loss function \eqref{eq:loss_AVIH} are introduced in the following sections. \cref{fig:avih} illustrates the schematic diagram of this method.

\paragraph*{Task loss ($L_t$)} 
The task loss is related to the preservation of the service model's output. With the service model $f_s(\cdot)$, the loss function is defined as follows:
\begin{equation}
    L_t(x_i, x_i') = \Vert f_s(x_i) - f_s(x_i') \Vert_2,
\end{equation}
where the distance metric is defined using $\ell_2$ norm; howeer, other distance metrics such as $\ell_1$ and $\ell_\infty$ norms can be also used. 

\paragraph*{Difference loss ($L_d$)}  
The difference loss is related to destroying the image's structural information. Motivated by Type-I adversarial attacks, this loss function is defined by the $\ell_1$ or $\ell_2$ distance between $x_i$ and $x_i'$ as follows:
\begin{equation}
    L_d(x_i,x_i') = \Vert x_i - x_i' \Vert_2. 
\end{equation}

\paragraph*{Recovery loss ($L_r$)}
Let us define the key model as $f_k(\cdot)$. To recover the original image, the recovery loss is defined by the $\ell_2$ distance between the original image $x_i$ and the recovered image $f_k(x_i')$ as follows:
\begin{equation}
    L_r(x_i,x_i') = \Vert x_i - f_k(x_i') \Vert_2.
\end{equation}

\paragraph*{Variance consistency loss ($L_v$)}
The variance consistency (VC) loss, proposed in \cite{su2023hiding}, addresses the trade-off between protection quality and recovery quality. 
To formulate this loss function, we divide the encrypted image $x_i'$ into $N$ overlapping patches, each with height $h$ and width $w$. 
Let us define the sum of the pixel values of the $n$-th patch as $S_n$, where $S_{n}^{(r)}$, $S_{n}^{(g)}$, and $S_{n}^{(b)}$ denote the sum of pixel values for the red, green, and blue channels, respectively. 
The VC loss then measures the channel-wise variance of $S_n$ as follows:
\begin{equation}
    L_v(x_i') = \mathtt{var}(S_{n}^{(r)})  + \mathtt{var}(S_{n}^{(g)})  + \mathtt{var}(S_{n}^{(b)}),
\end{equation}
where $\mathtt{var}(S_{n}^{(r)})$ denotes the empirical variance of the sequence $S_{1}^{(r)}$, $S_{2}^{(r)}$, $\cdots$, $S_{N}^{(r)}$. 
\begin{remark}
    If we aim to obtain an image that is robust against potential attackers, we should set a higher weight on the VC loss. Consequently, the quality of the image recovered using the key model will be poor.
\end{remark}

\subsection{Proposed Data Reconstruction Attack \label{subsec:proposed}}

Here, we first introduce the threat scenario of the AVIH method. 
Then, we present the proposed data reconstruction attack for the gallery dataset, where the visual information is hidden in noise-like images.

\begin{figure}[tb]
    \centering
    \includegraphics[width=\if 1\doublecolumn 0.99 \else 0.5 \fi\linewidth]{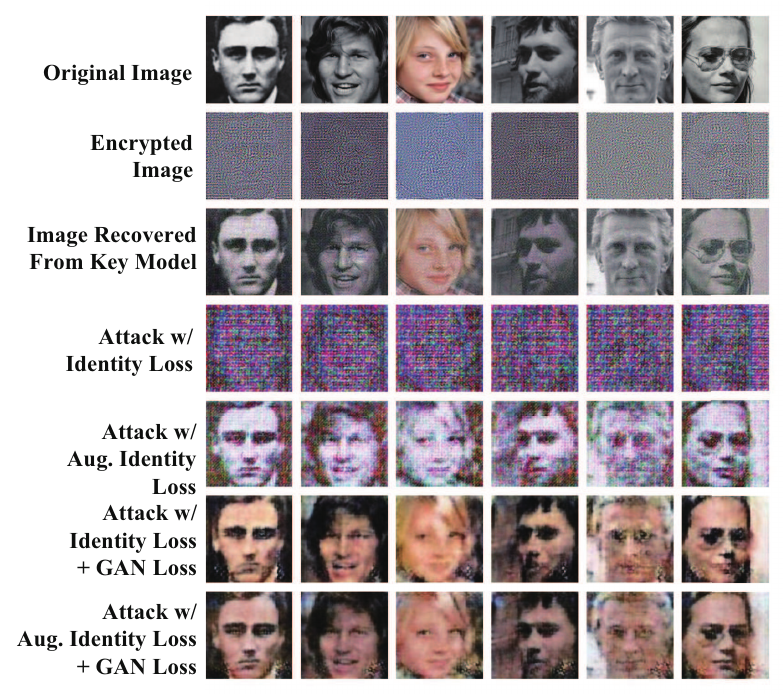}
    \caption{Ablation study for the key features of our proposed work (augmented identity loss and GAN loss). Here, we assume that 3\% of the encrypted data shares the same key model. }
    \label{fig:ablation_example}
\end{figure}

\paragraph*{Threat scenario}

In this work, we follow the system model in the AVIH method, where the target service type is image recognition. 
In an image recognition system, the class of a target image can be identified by comparing it with images in the gallery dataset. 
As shown in \cref{fig:avih}, the AVIH method transforms the images in the gallery dataset into noise-like images while preserving the output of the service DNN. 

Let us define the original gallery dataset as $\mathcal{G}$ and the encrypted gallery dataset as $\Tilde{\mathcal{G}}$. 
We consider a threat scenario where a malicious attacker aims to reconstruct the hidden images from noise-like images with access to the \textbf{1) encrypted dataset and 2) service DNN model}. 
According to \cite{su2023hiding}, the key model can reconstruct the original gallery dataset from the encrypted gallery dataset and is only available on the client-side.

More specifically, without making strong assumptions, since the service DNN weights are available at the remote server, we assume \textbf{a white-box access scenario}, \ie, an attacker can access the weights of the service DNN model.

\paragraph*{Motivation}

In this work, we focus on the fact that a private key model can reconstruct the original gallery dataset. 
This implies there is an \textbf{unknown but specific relationship} between the encrypted and original datasets. 
Motivated by this, we aim to mimic the functionality of the key model.
However, we neither have access to the key model nor know how it was trained. 
To address this, we leverage the generative model \cite{goodfellow2014generative}, which is widely used for guaranteeing photo-realism. 
However, generative model inversion attacks typically focus on mimicking the distribution of the training dataset and cannot reconstruct an image corresponding to a specific DNN output.
To directly reconstruct the original gallery dataset, we consider the attacker key model as $a_k(\cdot;\theta)$, where $\theta$ denotes its weights. For simplicity, we use $a_k(\cdot;\theta)$ and $a_k(\cdot)$ interchangeably.

\begin{figure*}
    \centering
    \includegraphics[width=0.85\textwidth]{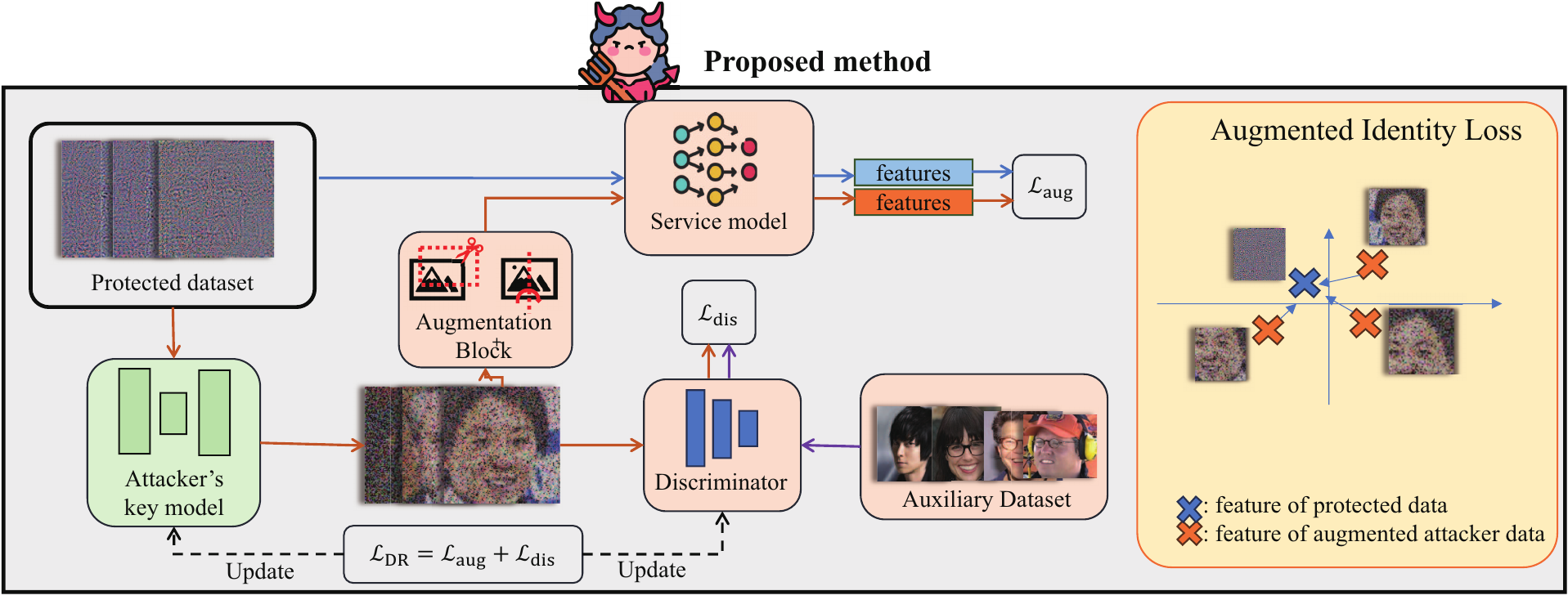}
    \caption{Illustration of the proposed data reconstruction attack for the AVIH method. The attacker (possibly an honest-but-curious cloud server manager) is assumed to have access to the protected dataset and the auxiliary dataset. The attacker first initializes a key model. Then, the key model's weights are updated to be more photo-realistic (discriminator loss, $\mathcal{L}_\text{dis}$) and to reconstruct the original image via feature matching (augmented identity loss, $\mathcal{L}_\text{aug}$). }
    \label{fig:proposed}
\end{figure*}

\paragraph*{Identity loss}
With the given information, we can assume that the target key model also preserves the output of the service model. 
Additionally, since we have access to the weights of the service DNN, we can formulate the identity loss as follows:
\begin{equation}\label{eq:identity_loss}
    L_{\text{I}}(\tilde{\mathcal{G}}) = \mathbb{E}_{x'\sim \tilde{\mathcal{G}}}\Vert f_s(a_k(x_i';\theta)) - f_s(x_i') \Vert_2,
\end{equation}
where $\tilde{\mathcal{G}}$ denotes the encrypted images sharing the same key model. Instead of an $\ell_2$-based loss, the identity loss function can be formulated to maximize the log-likelihood of the target class $c$ or the cosine similarity of two feature vectors.

\paragraph*{Overfitting in data reconstruction}

In \cref{fig:ablation_example}, we show examples of the original images and the encrypted images by the AVIH method. 
The figure also depicts the images recovered by the original key model and the attacker key model. As shown, the attacker model trained with identity loss in \eqref{eq:identity_loss} recovers very noisy images compared to the original images or those recovered by the original key model. 
This issue is quite similar to \textbf{overfitting} in ordinary machine learning problems. More specifically, there is a true relationship between the encrypted images and the original images, represented by the original key model. However, a few encrypted images are not sufficient to demonstrate this relationship using the service model. 
Therefore, in the remainder of this section, we propose 1) augmented identity loss and 2) a GAN-based training scheme to secure the generalization of the trained attacker key model. The details of these methods are depicted in \cref{fig:proposed}.

\paragraph*{Augmented identity loss}

In typical machine learning model training, data augmentation is widely used to increase the validation/test accuracy of the trained model, \ie, for better generalization
Similarly, to alleviate the overfitting issue in data reconstruction, we combine data augmentation with identity loss.
Let us consider an image $x'$ drawn from the encrypted dataset $\tilde{\mathcal{G}}$. Then, we may reconstruct the original image using $a_k(x';\theta)$. Unlike the canonical identity loss in \eqref{eq:identity_loss}, we apply random data augmentation before forwarding the image into the service model $f_x(\cdot)$. Denoting the random augmentation process as $T(\cdot)$, the identity loss in \eqref{eq:identity_loss} can be redefined as follows:
\begin{equation}\label{eq:aug_id_loss}
        L_{\text{I}}(\tilde{\mathcal{G}}) = \mathbb{E}_{x'\sim \tilde{\mathcal{G}}}\Vert f_s(T(a_k(x_i';\theta))) - f_s(x_i') \Vert_2.
\end{equation}

In our work, we consider the following data augmentation methods: i) random horizontal flip, ii) random padding, and iii) random crop. In the right part of \cref{fig:proposed}, we illustrate the concept of the augmented identity loss.
For further generalization, we also experimented with randomized smoothing on the reconstructed data $a_k(x';\theta)$; however, it did not produce notable differences.

\paragraph*{Generative model inversion attack}

In this paragraph, we aim to resolve the overfitting issue using a GAN-based loss function. 
Intuitively, if we want to find an image that has the same output as the encrypted image $x'$, there would be many possible images, most of which are unnatural. By reducing the number of cases by restricting the unnatural images, we can resolve the overfitting issue.

To make the reconstructed images natural, we consider an optimization problem that minimizes the augmented identity loss with a Jensen-Shannon (JS) divergence constraint between the auxiliary dataset and the reconstructed images as follows:
\begin{equation}\label{eq:opt_problem}
    \begin{split}
        \min_{a_k(\cdot)}~~~ L_\mathrm{I}(\tilde{\mathcal{G}}), ~~~\text{s.t.}~~~ D_\text{JS}(a_k(\tilde{\mathcal{G}})\Vert \mathcal{X}) \le \epsilon,
    \end{split}
\end{equation}
where $\mathcal{X}$ denotes the auxiliary dataset, and $D_{\text{JS}}(a_k(\tilde{\mathcal{G}})\Vert \mathcal{X})$ denotes the JS divergence between the reconstructed images $a_k(x'), x'\sim \tilde{\mathcal{G}}$ and the auxiliary images $x''\sim\mathcal{X}$. 
From this optimization, we derive the Lagrangian of the problem \eqref{eq:opt_problem} as follows:
\begin{equation}\label{eq:Lagrangian}
    \mathcal{L} = L_\mathrm{I}(\tilde{\mathcal{G}}) + \lambda (D_\text{JS} (a_k(\tilde{\mathcal{G}})\Vert \mathcal{X}) - \epsilon),
\end{equation}
where $\mu \ge 0$. 
Let us consider the minimizer $a_k(\cdot;\theta)$ of the problem as $a^*_k(\cdot;\theta)$. Then, for any constant $\epsilon$, there exists $\mu \ge 0$ that makes $a^*_k(\cdot;\theta)$ a minimizer of the original problem \eqref{eq:opt_problem}. The intuition is that for the minimizer $a^*_k(\cdot;\theta)$, a larger value of $\lambda$ indicates a smaller $\epsilon$ in \eqref{eq:opt_problem}. 
Hence, we aim to minimize the Lagrangian in \eqref{eq:Lagrangian}.

The next step is to convert the function in \eqref{eq:Lagrangian} into a GAN formulation. As shown in \cite{goodfellow2014generative}, the JS divergence minimization problem can be replaced by the GAN optimization problem. 
For brevity, we use simpler notation: the probability density function of the recovered images $x^* \sim a_k(\tilde{\mathcal{G}}) = p(x)$ and the auxiliary images $x'' \sim \mathcal{X} = q(x)$. The JS divergence in \eqref{eq:Lagrangian} can then be rewritten as follows:
\if 1\doublecolumn
\begin{equation}
    \begin{split}
        &  D_\text{JS}(a_k(\tilde{\mathcal{G}}) \Vert \mathcal{X}))  = D_\text{JS}(p \Vert q )\\
        & \propto \underset{x^* \sim p(x)}{\mathbb{E}} \left[ \log \frac{2 p(x^*)}{p(x^*) + q(x^*)} \right] \\ 
        & ~~ + \underset{x'' \sim q(x)}{\mathbb{E}} \left[ \log \frac{2 q(x'')}{p(x'') + q(x'')} \right].
    \end{split}
\end{equation}
\else
\begin{equation}
    \begin{split}
        &  D_\text{JS}(a_k(\tilde{\mathcal{G}}) \Vert \mathcal{X}))  =  D_\text{JS}(p \Vert q ) \\ 
        & \propto \underset{x^* \sim p(x)}{\mathbb{E}} \left[ \log \frac{2 p(x^*)}{p(x^*) + q(x^*)} \right]  + \underset{x'' \sim q(x)}{\mathbb{E}} \left[ \log \frac{2 q(x'')}{p(x'') + q(x'')} \right].
    \end{split}
\end{equation}
\fi

By defining a discriminator as $D(x)$, we can convert the JS divergence into a GAN formulation as follows:
\if 1\doublecolumn
\begin{equation}
    \begin{split}
        &D_\text{JS}(p\Vert q) = \\ 
        &\max_D \mathbb{E}_{x^*\sim p}\left[\log(D(x^*))\right] + \mathbb{E}_{x''\sim q}\left[\log(1- D(x''))\right],
    \end{split}
\end{equation}
\else
\begin{equation}
    \begin{split}
        D_\text{JS}(p\Vert q) = \max_D \mathbb{E}_{x^*\sim p}\left[\log(D(x^*))\right] + \mathbb{E}_{x''\sim q}\left[\log(1- D(x''))\right],
    \end{split}
\end{equation}
\fi
where the optimal $D$ is $\frac{p(x)}{p(x) + q(x)}$.
Then, the loss function for the attacker key model is defined by 
\begin{equation}
    L_{\text{key}}(\tilde{\mathcal{G}}) = \mathbb{E}_{x'\sim \tilde{\mathcal{G}}} \left[ \log D(a_k(x';\theta)) \right] + \lambda_1 \cdot L_{\text{I}}(\tilde{\mathcal{G}}),
\end{equation}
where $D$ is the discriminator model, and the last layer is activated by hyperbolic tangent function. 
Similarly, the discriminator loss is defined as follows:
\if 1\doublecolumn
\begin{multline}
    L_{\text{dis}}(\tilde{\mathcal{G}}) = - \mathbb{E}_{x'\sim \tilde{\mathcal{G}}}\left[\log D(a_k(x';\theta))\right] \\ 
    - \mathbb{E}_{x\sim \mathcal{D}_{\text{aux}}}\left[\log (1-D(x))\right].
\end{multline}
\else
\begin{equation}
    L_{\text{dis}}(\tilde{\mathcal{G}}) = - \mathbb{E}_{x'\sim \tilde{\mathcal{G}}}\left[\log D(a_k(x';\theta))\right] 
    - \mathbb{E}_{x\sim \mathcal{D}_{\text{aux}}}\left[\log (1-D(x))\right].
\end{equation}
\fi

Since the original images and the auxiliary images are not identical but belong to the same category (e.g., face images), we use a patch-GAN model for our optimization, where the discriminator $D$ classifies true and false patches of the images. In \cref{fig:proposed}, the attacker aims to generate photo-realistic results by deceiving the discriminator.
\begin{remark}[JS divergence vs. KL divergence]
    In our problem formulation \eqref{eq:opt_problem}, we use the JS divergence between the recovered images and auxiliary images as our constraint for photo-realism. Another metric, KL divergence, is widely used to ensure similarity between two datasets. We have tried with the KL divergence formulation; however, it is closely related to variational inference, which requires a pre-trained GAN. Since a pre-trained GAN is not suitable for exact data reconstruction attacks, we use JS divergence as our constraint.
\end{remark}

\subsection{Ablation Study of Key Features}

Before introducing our experimental results, we briefly present graphical examples of our attacker key model with and without our key features for resolving overfitting issues. In the previous subsection, we proposed the augmented identity loss and GAN-based training loss. 
In \cref{fig:ablation_example}, we show our results on the AgeDB-30 dataset, using the CelebA dataset as the auxiliary dataset. As depicted in the fourth column, the results with the canonical identity loss recover images that are not very similar to the original ones. However, by leveraging augmented identity loss, the shape of the images can be recovered, though the colors are not realistic. On the other hand, using GAN-based training yields more photo-realistic recovered images. Moreover, by combining both methods, the quality of the recovered images is further enhanced.

\section{Experiments}
\label{sec:experiments}
In this section, we evaluate the proposed data reconstruction attack against adversarial visual information hiding. Since there has been no prior work on exact data reconstruction attack methods for deep neural network models, we measure the quality of the reconstruction using various metrics\footnote{Only the authors of \cite{haim2022reconstructing} have shown exact data reconstruction, but their method requires the target DNN to be a homogeneous neural network, which is not a practical assumption.}. 
Instead of comparing with other methods, we conduct an ablation study on our key contributions: 1) augmented identity loss and 2) GAN-based training.

\subsection{Experimental Details}

We conduct two main experiments: one for the face recognition scenario and another for the object re-identification scenario. Both experiments are performed on a workstation equipped with an AMD Ryzen R9 5950x 16-core CPU and an NVIDIA GeForce RTX 3090 GPU with 24GB of VRAM.

\paragraph*{Face recognition scenario}

In the face recognition scenario, we perform data reconstruction attacks on three target face datasets: LFW \cite{LFWTech}, AgeDB-30 \cite{moschoglou2017agedb}, and CFP-FP \cite{cfp-paper}. For the service model on the local server and cloud, we use ArcFace \cite{deng2019arcface} and AdaFace \cite{kim2022adaface} models with IR-18 and IR-50 backbones. For the data reconstruction attack, we use the Celeb-A \cite{liu2015faceattributes} dataset as our auxiliary dataset for all target face datasets.
As an evaluation metric for face recognition, we use TAR@FAR=0.01 for the reconstructed images via our method. For this accuracy evaluation, we use the AdaFace model with an IR-101 backbone.

\paragraph*{Implementation of AVIH}

For the implementation of the AVIH method, we follow the hyperparameters used in the original paper~\cite{su2023hiding}. For example, we update encrypted images ($x'$) for 800 epochs, and the kernel size for the VC loss is set to 4. The weights for the difference loss, recovery loss, and VC loss are set to 0.03, 0.5, and 3.0, respectively. The key model is configured using the standard U-Net~\cite{Unet}.
In our experiment, we slightly modify the gallery sets of the datasets to contain 2,000 images. We then run the AVIH method on these 2,000 gallery images, using 1,000 images for training our attacker key model and the remaining 1,000 images for evaluating the trained attacker key model.

\paragraph*{Implementation of our data reconstruction}

In our data reconstruction attack, we train an attacker key model based on the U-Net structure\footnote{We tried other structures by modifying the U-Net structure, but the results were almost the same.}. We trained our key model for 1,600 steps with a batch size of 32. The weight on the augmented identity loss is set to 30.0. 
For the augmented identity loss function, we use the following data augmentation methods: random horizontal flip, random padding of 5 pixels, and random cropping to the original size.

\begin{table}[tb]
    \caption{
    The accuracy measurements of the original gallery images, protected images, reconstructed images by original key model, and our data reconstruction attack results. 
    }
    \label{tab:accuracy_1}
    \centering
    \adjustbox{width= \if 1\doublecolumn .85 \else 0.5 \fi \linewidth }{
    \begin{tabular}{ccccc}
    \toprule
         & \multirow{2}{*}{\makecell{Same-key \\ data (\%)}} & \multicolumn{3}{c}{TPR $\uparrow$} \\
    \cmidrule{3-5}
         &  & AgeDB-30 & LFW & CFP-FP \\
    \midrule
        \multirow{4}{*}{Ours} 
        & 1\% & 0.628\textsuperscript{$\pm$0.058} & 0.612\textsuperscript{$\pm$0.136} & 0.554\textsuperscript{$\pm$0.054} \\
        & 3\% & 0.746\textsuperscript{$\pm$0.043} & 0.825\textsuperscript{$\pm$0.039} & 0.630\textsuperscript{$\pm$0.018}\\
        & 10\% & 0.761\textsuperscript{$\pm$0.254} & 0.737\textsuperscript{$\pm$0.341} & 0.663\textsuperscript{$\pm$0.039} \\
        & 70\% & \textbf{0.924\textsuperscript{$\pm$0.011}} & \textbf{0.970\textsuperscript{$\pm$0.006}} & \textbf{0.817\textsuperscript{$\pm$0.019}} \\
    \cmidrule{1-5}
        Original & - & 0.980\textsuperscript{$\pm$0.000} & 0.998\textsuperscript{$\pm$0.000} & 0.971\textsuperscript{$\pm$0.000} \\
        Protected & - & 0.211\textsuperscript{$\pm$0.000} & 0.149\textsuperscript{$\pm$0.000} & 0.579\textsuperscript{$\pm$0.000}\\
        Key model & - & 0.971\textsuperscript{$\pm$0.000} & 0.998\textsuperscript{$\pm$0.000}  & 0.964\textsuperscript{$\pm$0.000} \\
        Random  & - & 0.215\textsuperscript{$\pm$0.000} & 0.036\textsuperscript{$\pm$0.000} &  0.590\textsuperscript{$\pm$0.000} \\
    \bottomrule
    \end{tabular}
    }
    \setlength\tabcolsep{5pt} %
\end{table}

\paragraph*{Evaluation metrics for reconstruction quality measurement}
To measure the quality of the reconstructed images from our data reconstruction attack, we used the following evaluation metrics: 1) mean square error (MSE), 2) learned perceptual image patch similarity (LPIPS)~\cite{zhang2018perceptual}, 3) peak signal-to-noise ratio (PSNR), 4) contrastive language-image pre-training (CLIP)~\cite{hessel2021clipscore}, and 5) structural similarity index measure (SSIM)~\cite{1284395}.

\subsection{Accuracy and Similarity Metrics }

In our experiments, before implementing our data reconstruction attack scheme, the AVIH~\cite{su2023hiding} encrypts the gallery dataset of three face recognition datasets: AgeDB-30, LFW, and CFP-FP. We note that the encrypted gallery dataset successfully performs face recognition tasks for the target service model. For example, with the encrypted LFW gallery dataset, the cloud achieves a TPR accuracy of 98.40\%, which is the same accuracy as with the original gallery dataset.

\paragraph*{Accuracy metrics}

In \cref{tab:accuracy_1}, we present the TPR performance metrics at FPR 0.01 across three datasets: AgeDB-30, LFW, and CFP-FP. The TPR accuracy values are provided for the proposed method (\textbf{Ours}) at different percentages of images sharing the same key model (1\%, 3\%, 10\%, and 70\%). We also compare the results of our method with the following: 1) original gallery dataset, 2) encrypted dataset, 3) dataset reconstructed by the key model, and 4) random face images.

For all three datasets, TPR values of the proposed method gradually increase as more gallery images share the same key model. For instance, in the AgeDB-30 dataset, the proposed method achieves a TPR of 0.628 with 1\% same-key data, which increases to 0.746 with 3\%, 0.761 with 10\%, and 0.924 with 70\% same-key data.

On the other hand, with the original gallery datasets, the TPR accuracy values are sufficient for recognizing most of the query images. For example, the TPR value for the AgeDB-30 dataset is 98.0\%. More importantly, the gallery dataset reconstructed by the original private key model performs almost the same as the original gallery dataset. The encrypted gallery set has significantly lower TPR values since the evaluation service model (IR-101 backbone) is different from the target service model (IR-18 backbone).

To summarize, the proposed method shows significant improvement in TPR by executing data reconstruction attacks against the AVIH encryption method.

\begin{table}[tb]
    \centering
    \caption{
    The quality measurements of the original gallery images, protected images, reconstructed images by original key model, and our data reconstruction attack results. The dataset is AgeDB-30 dataset, and the used backbone network model is the IR-18 network model. 
    }
    \label{tab:image_sim_1}
    \adjustbox{width=\if 1\doublecolumn .99 \else 0.7 \fi\linewidth}{
    \begin{tabular}{c|c|ccccc}
    \toprule
         & & MSE $\downarrow$ & LPIPS $\downarrow$ & PSNR $\uparrow$ & CLIP $\uparrow$ & SSIM $\uparrow$ \\
    \midrule
        \parbox[t]{2mm}{\multirow{7}{*}{\rotatebox[origin=c]{90}{AgeDB-30 Dataset}}} & Ours (1\%) & 0.114\textsuperscript{$\pm$0.007} & 0.540\textsuperscript{$\pm$0.008} & 9.960\textsuperscript{$\pm$0.268} & 0.715\textsuperscript{$\pm$0.006} & 0.249\textsuperscript{$\pm$0.003} \\
         & Ours (3\%) & 0.071\textsuperscript{$\pm$0.006} & 0.450\textsuperscript{$\pm$0.018} & 11.847\textsuperscript{$\pm$0.394} & 0.746\textsuperscript{$\pm$0.013} & 0.270\textsuperscript{$\pm$0.005} \\
         & Ours (10\%) & 0.121\textsuperscript{$\pm$0.149} & 0.401\textsuperscript{$\pm$0.069} & 10.996\textsuperscript{$\pm$2.766} & 0.774\textsuperscript{$\pm$0.017} & 0.269\textsuperscript{$\pm$0.046} \\
         & Ours (70\%) & \textbf{0.063\textsuperscript{$\pm$0.008}} & \textbf{0.305\textsuperscript{$\pm$0.017}} & \textbf{12.464\textsuperscript{$\pm$0.593}} & \textbf{0.811\textsuperscript{$\pm$0.009}} & \textbf{0.298\textsuperscript{$\pm$0.006}} \\
    \cmidrule{2-7}
        & Original  & 0.001\textsuperscript{$\pm$0.000} & 0.009\textsuperscript{$\pm$0.000} & 29.563\textsuperscript{$\pm$0.000} & 0.966\textsuperscript{$\pm$0.000} & 0.461\textsuperscript{$\pm$0.000} \\
        & Protected  & 0.662\textsuperscript{$\pm$0.000} & 1.214\textsuperscript{$\pm$0.000} & 1.866\textsuperscript{$\pm$0.000} & 0.641\textsuperscript{$\pm$0.000} & 0.019\textsuperscript{$\pm$0.000} \\
        & Key model  & 0.022\textsuperscript{$\pm$0.000} & 0.367\textsuperscript{$\pm$0.000} & 17.525\textsuperscript{$\pm$0.000} & 0.837\textsuperscript{$\pm$0.000} & 0.302\textsuperscript{$\pm$0.000} \\
        \midrule
        \parbox[t]{2mm}{\multirow{7}{*}{\rotatebox[origin=c]{90}{LFW Dataset}}} & Ours (1\%) & 0.142\textsuperscript{$\pm$0.077} & 0.527\textsuperscript{$\pm$0.043} & 9.120\textsuperscript{$\pm$1.781} & 0.747\textsuperscript{$\pm$0.022} & 0.246\textsuperscript{$\pm$0.026} \\
         & Ours (3\%) & 0.060\textsuperscript{$\pm$0.007} & 0.377\textsuperscript{$\pm$0.009} & 12.395\textsuperscript{$\pm$0.492} & 0.813\textsuperscript{$\pm$0.017} & 0.291\textsuperscript{$\pm$0.004} \\
         & Ours (10\%) & 0.140\textsuperscript{$\pm$0.181} & 0.352\textsuperscript{$\pm$0.086} & 11.232\textsuperscript{$\pm$3.934} & 0.832\textsuperscript{$\pm$0.042} & 0.276\textsuperscript{$\pm$0.061} \\
         & Ours (70\%) & \textbf{0.048\textsuperscript{$\pm$0.008}} & \textbf{0.268\textsuperscript{$\pm$0.014}} & \textbf{13.366\textsuperscript{$\pm$0.673}} & \textbf{0.865\textsuperscript{$\pm$0.011}} & \textbf{0.323\textsuperscript{$\pm$0.006}} \\
    \cmidrule{2-7}
        & Original  & 0.001\textsuperscript{$\pm$0.000} & 0.008\textsuperscript{$\pm$0.000} & 32.067\textsuperscript{$\pm$0.000} & 0.981\textsuperscript{$\pm$0.000} & 0.477\textsuperscript{$\pm$0.000} \\
        & Protected  & 0.622\textsuperscript{$\pm$0.000} & 1.302\textsuperscript{$\pm$0.000} & 2.108\textsuperscript{$\pm$0.000} & 0.556\textsuperscript{$\pm$0.000} & 0.015\textsuperscript{$\pm$0.000} \\
        & Key model  & 0.006\textsuperscript{$\pm$0.000} & 0.243\textsuperscript{$\pm$0.000} & 23.071\textsuperscript{$\pm$0.000} & 0.919\textsuperscript{$\pm$0.000} & 0.385\textsuperscript{$\pm$0.000} \\
            \midrule
        \parbox[t]{2mm}{\multirow{7}{*}{\rotatebox[origin=c]{90}{CFP-FP Dataset}}} & Ours (1\%) & 0.210\textsuperscript{$\pm$0.020} & 0.623\textsuperscript{$\pm$0.019} & 7.136\textsuperscript{$\pm$0.419} & 0.728\textsuperscript{$\pm$0.011} & 0.163\textsuperscript{$\pm$0.003} \\
         & Ours (3\%) & 0.214\textsuperscript{$\pm$0.047} & 0.566\textsuperscript{$\pm$0.027} & 7.272\textsuperscript{$\pm$0.891} & 0.755\textsuperscript{$\pm$0.015} & 0.177\textsuperscript{$\pm$0.004} \\
         & Ours (10\%) & \textbf{0.176\textsuperscript{$\pm$0.030}} & 0.456\textsuperscript{$\pm$0.018} & \textbf{8.007\textsuperscript{$\pm$0.676}} & 0.804\textsuperscript{$\pm$0.010} & 0.187\textsuperscript{$\pm$0.004} \\
         & Ours (70\%) & 0.210\textsuperscript{$\pm$0.053} & \textbf{0.408\textsuperscript{$\pm$0.015}} & 7.310\textsuperscript{$\pm$0.982} & \textbf{0.835\textsuperscript{$\pm$0.005}} & \textbf{0.204\textsuperscript{$\pm$0.004}} \\
    \cmidrule{2-7}
        & Original  & 0.001\textsuperscript{$\pm$0.000} & 0.009\textsuperscript{$\pm$0.000} & 29.563\textsuperscript{$\pm$0.000} & 0.966\textsuperscript{$\pm$0.000} & 0.461\textsuperscript{$\pm$0.000} \\
        & Protected  & 0.636\textsuperscript{$\pm$0.000} & 1.219\textsuperscript{$\pm$0.000} & 2.044\textsuperscript{$\pm$0.000} & 0.633\textsuperscript{$\pm$0.000} & 0.020\textsuperscript{$\pm$0.000} \\
        & Key model  & 0.028\textsuperscript{$\pm$0.000} & 0.435\textsuperscript{$\pm$0.000} & 16.377\textsuperscript{$\pm$0.000} & 0.825\textsuperscript{$\pm$0.000} & 0.285\textsuperscript{$\pm$0.000} \\
    \bottomrule
    \end{tabular}
    }
    \setlength\tabcolsep{5pt} %
\end{table}

\begin{table}[tb]
    \setlength\tabcolsep{4pt} %
    \centering
    \caption{
    The accuracy and quality measurements of the original gallery images, protected images, reconstructed images by original key model, and our data reconstruction attack results. The used dataset is AgeDB-30, and the target face recognition model is AdaFace and ArcFace with the IR-50 backbone netowrk model. 
    }
    \label{tab:image_backbone_model}
    \adjustbox{width=\if 1\doublecolumn .99 \else 0.7 \fi\linewidth}{
    \begin{tabular}{c|c|ccccc}
    \toprule
         & & TPR $\uparrow$ & LPIPS $\downarrow$ & PSNR $\uparrow$ & CLIP $\uparrow$ & SSIM $\uparrow$ \\
    \midrule
        \parbox[t]{2mm}{\multirow{7}{*}{\rotatebox[origin=c]{90}{AdaFace \& IR-50}}} & Ours (1\%) & 0.719\textsuperscript{$\pm$0.045} & 0.530\textsuperscript{$\pm$0.012} & 10.391\textsuperscript{$\pm$0.377} & 0.719\textsuperscript{$\pm$0.007} & 0.256\textsuperscript{$\pm$0.004} \\
         & Ours (3\%) & 0.861\textsuperscript{$\pm$0.031} & 0.433\textsuperscript{$\pm$0.017} & 12.331\textsuperscript{$\pm$0.418} & 0.747\textsuperscript{$\pm$0.007} & 0.284\textsuperscript{$\pm$0.007} \\
         & Ours (10\%) & 0.898\textsuperscript{$\pm$0.020} & 0.378\textsuperscript{$\pm$0.016} & 12.594\textsuperscript{$\pm$0.630} & 0.772\textsuperscript{$\pm$0.011} & 0.303\textsuperscript{$\pm$0.005} \\
         & Ours (70\%) & \textbf{0.932\textsuperscript{$\pm$0.011}} & \textbf{0.341\textsuperscript{$\pm$0.027}} & \textbf{12.599\textsuperscript{$\pm$0.973}} & \textbf{0.797\textsuperscript{$\pm$0.013}} & \textbf{0.310\textsuperscript{$\pm$0.007}} \\
    \cmidrule{2-7}
        & Original  & 0.980\textsuperscript{$\pm$0.000} & 0.009\textsuperscript{$\pm$0.000} & 29.563\textsuperscript{$\pm$0.000} & 0.966\textsuperscript{$\pm$0.000} & 0.461\textsuperscript{$\pm$0.000} \\
        & Protected  & 0.311\textsuperscript{$\pm$0.000} & 1.219\textsuperscript{$\pm$0.000} & 2.044\textsuperscript{$\pm$0.000} & 0.633\textsuperscript{$\pm$0.000} & 0.020\textsuperscript{$\pm$0.000} \\
        & Key model  & 0.952\textsuperscript{$\pm$0.000} & 0.435\textsuperscript{$\pm$0.000} & 16.377\textsuperscript{$\pm$0.000} & 0.825\textsuperscript{$\pm$0.000} & 0.285\textsuperscript{$\pm$0.000} \\
        \midrule
        \parbox[t]{2mm}{\multirow{7}{*}{\rotatebox[origin=c]{90}{ArcFace \& IR-50}}} & Ours (1\%) & 0.709\textsuperscript{$\pm$0.044} & 0.537\textsuperscript{$\pm$0.010} & 10.348\textsuperscript{$\pm$0.233} & 0.716\textsuperscript{$\pm$0.007} & 0.253\textsuperscript{$\pm$0.004} \\
         & Ours (3\%) & 0.823\textsuperscript{$\pm$0.020} & 0.440\textsuperscript{$\pm$0.015} & 11.787\textsuperscript{$\pm$0.432} & 0.753\textsuperscript{$\pm$0.008} & 0.278\textsuperscript{$\pm$0.009} \\
         & Ours (10\%) & 0.885\textsuperscript{$\pm$0.012} & 0.373\textsuperscript{$\pm$0.016} & 12.122\textsuperscript{$\pm$0.599} & 0.776\textsuperscript{$\pm$0.008} & 0.291\textsuperscript{$\pm$0.004} \\
         & Ours (70\%) & \textbf{0.917\textsuperscript{$\pm$0.012}} & \textbf{0.333\textsuperscript{$\pm$0.022}} & \textbf{12.338\textsuperscript{$\pm$0.377}} & \textbf{0.802\textsuperscript{$\pm$0.007}} & \textbf{0.304\textsuperscript{$\pm$0.004}} \\
    \cmidrule{2-7}
        & Original  & 0.980\textsuperscript{$\pm$0.000} & 0.009\textsuperscript{$\pm$0.000} & 29.563\textsuperscript{$\pm$0.000} & 0.966\textsuperscript{$\pm$0.000} & 0.461\textsuperscript{$\pm$0.000} \\
        & Protected  & 0.167\textsuperscript{$\pm$0.000} & 1.160\textsuperscript{$\pm$0.000} & 2.104\textsuperscript{$\pm$0.000} & 0.603\textsuperscript{$\pm$0.000} & 0.020\textsuperscript{$\pm$0.000} \\
        & Key model  & 0.982\textsuperscript{$\pm$0.000} & 0.222\textsuperscript{$\pm$0.000} & 20.664\textsuperscript{$\pm$0.000} & 0.868\textsuperscript{$\pm$0.000} & 0.351\textsuperscript{$\pm$0.000} \\
    \bottomrule
    \end{tabular}
    }
\end{table}

\begin{table*}[tb]
    \centering
    \caption{Ablation study for our key contributions: 1) GAN-based training and 2) Augmented identification (ID) loss. In this table, the target dataset is AgeDB-30 dataset, where the target service model is AdaFace with the IR-18 backbone network. 
    As presented in this table, our key contributions effectively resolve the overfitting issues on the data reconstruction when a small portion of the gallery images share the same key model. 
    }
    \adjustbox{width=0.8\linewidth}{
    \begin{tabular}{ccccccccccc}
    \toprule
        \# Leaked data & GAN Loss & Aug. ID. Loss & TPR $\uparrow$ & MSE $\downarrow$ & LPIPS $\downarrow$ & PSNR $\uparrow$ & CLIP $\uparrow$ & SSIM $\uparrow$ \\
    \midrule
        \multirow{4}{*}{1\%} & \xmarkg & \xmarkg & 0.329\textsuperscript{$\pm$0.110} & 0.418\textsuperscript{$\pm$0.050} & 0.945\textsuperscript{$\pm$0.073} & 3.954\textsuperscript{$\pm$0.512} & 0.653\textsuperscript{$\pm$0.072} & 0.051\textsuperscript{$\pm$0.005} \\
        & \xmarkg & \cmark & 0.524\textsuperscript{$\pm$0.028} & 0.271\textsuperscript{$\pm$0.057} & 0.923\textsuperscript{$\pm$0.038} & 6.036\textsuperscript{$\pm$0.904} & \textbf{0.772\textsuperscript{$\pm$0.008}} & 0.094\textsuperscript{$\pm$0.009} \\
        & \cmark & \xmarkg & 0.517\textsuperscript{$\pm$0.091} & 0.116\textsuperscript{$\pm$0.013} & 0.571\textsuperscript{$\pm$0.015} & 9.840\textsuperscript{$\pm$0.508} & 0.709\textsuperscript{$\pm$0.005} & 0.236\textsuperscript{$\pm$0.008} \\
        & \cmark & \cmark & \textbf{0.628\textsuperscript{$\pm$0.058}} & \textbf{0.114\textsuperscript{$\pm$0.007}} & \textbf{0.540\textsuperscript{$\pm$0.008}} & \textbf{9.960\textsuperscript{$\pm$0.268}} & 0.715\textsuperscript{$\pm$0.006} & \textbf{0.249\textsuperscript{$\pm$0.003}} \\
    \midrule
        \multirow{4}{*}{3\%} & \xmarkg & \xmarkg & 0.299\textsuperscript{$\pm$0.134} & 0.437\textsuperscript{$\pm$0.060} & 0.937\textsuperscript{$\pm$0.109} & 3.792\textsuperscript{$\pm$0.600} & 0.657\textsuperscript{$\pm$0.057} & 0.052\textsuperscript{$\pm$0.004} \\
        & \xmarkg & \cmark & 0.549\textsuperscript{$\pm$0.051} & 0.323\textsuperscript{$\pm$0.060} & 0.897\textsuperscript{$\pm$0.038} & 5.318\textsuperscript{$\pm$0.759} & \textbf{0.781\textsuperscript{$\pm$0.010}} & 0.096\textsuperscript{$\pm$0.010} \\
        & \cmark & \xmarkg &  0.543\textsuperscript{$\pm$0.186} & 0.130\textsuperscript{$\pm$0.135} & 0.519\textsuperscript{$\pm$0.042} & 10.346\textsuperscript{$\pm$2.388} & 0.719\textsuperscript{$\pm$0.010} & 0.244\textsuperscript{$\pm$0.036} \\
        & \cmark & \cmark & \textbf{0.746\textsuperscript{$\pm$0.043}} & \textbf{0.071\textsuperscript{$\pm$0.006}} & \textbf{0.450\textsuperscript{$\pm$0.018}} & \textbf{11.847\textsuperscript{$\pm$0.394}} & 0.746\textsuperscript{$\pm$0.013} & \textbf{0.270\textsuperscript{$\pm$0.005}} \\
    \midrule    
        \multirow{4}{*}{10\%} & \xmarkg & \xmarkg & 0.370\textsuperscript{$\pm$0.084} & 0.375\textsuperscript{$\pm$0.034} & 0.999\textsuperscript{$\pm$0.054} & 4.419\textsuperscript{$\pm$0.379} & 0.615\textsuperscript{$\pm$0.041} & 0.058\textsuperscript{$\pm$0.004} \\
        & \xmarkg & \cmark & 0.662\textsuperscript{$\pm$0.063} & 0.264\textsuperscript{$\pm$0.039} & 0.700\textsuperscript{$\pm$0.036} & 6.244\textsuperscript{$\pm$0.641} & \textbf{0.807\textsuperscript{$\pm$0.007}} & 0.126\textsuperscript{$\pm$0.009} \\
        & \cmark & \xmarkg & \textbf{0.787\textsuperscript{$\pm$0.063}} & \textbf{0.073\textsuperscript{$\pm$0.010}} & 0.413\textsuperscript{$\pm$0.026} & \textbf{11.832\textsuperscript{$\pm$0.579}} & 0.761\textsuperscript{$\pm$0.014} & \textbf{0.272\textsuperscript{$\pm$0.013}} \\
        & \cmark & \cmark & 0.761\textsuperscript{$\pm$0.254} & 0.121\textsuperscript{$\pm$0.149} & \textbf{0.401\textsuperscript{$\pm$0.069}} & 10.996\textsuperscript{$\pm$2.766} & 0.774\textsuperscript{$\pm$0.017} & 0.269\textsuperscript{$\pm$0.046} \\
    \midrule
        \multirow{4}{*}{70\%} & \xmarkg & \xmarkg & 0.860\textsuperscript{$\pm$0.022} & 0.193\textsuperscript{$\pm$0.049} & 0.716\textsuperscript{$\pm$0.046} & 7.431\textsuperscript{$\pm$1.020} & 0.802\textsuperscript{$\pm$0.011} & 0.119\textsuperscript{$\pm$0.013} \\
        & \xmarkg & \cmark & 0.891\textsuperscript{$\pm$0.021} & 0.187\textsuperscript{$\pm$0.042} & 0.490\textsuperscript{$\pm$0.035} & 7.581\textsuperscript{$\pm$1.040} & \textbf{0.832\textsuperscript{$\pm$0.005}} & 0.197\textsuperscript{$\pm$0.013} \\
        & \cmark & \xmarkg & \textbf{0.929\textsuperscript{$\pm$0.009}} & 0.068\textsuperscript{$\pm$0.010} & 0.341\textsuperscript{$\pm$0.027} & 12.153\textsuperscript{$\pm$0.659} & 0.796\textsuperscript{$\pm$0.008} & \textbf{0.304\textsuperscript{$\pm$0.007}} \\
        & \cmark & \cmark & 0.924\textsuperscript{$\pm$0.011} & \textbf{0.063\textsuperscript{$\pm$0.008}} & \textbf{0.305\textsuperscript{$\pm$0.017}} & \textbf{12.464\textsuperscript{$\pm$0.593}} & 0.811\textsuperscript{$\pm$0.009} & 0.298\textsuperscript{$\pm$0.006} \\
    \bottomrule
    \end{tabular}
    }
    \setlength\tabcolsep{5pt} %
    \label{tab:main_face_table_2}
\end{table*}

\paragraph*{Image similarity metrics}

In \cref{tab:image_sim_1}, we present the evaluation results for the three face recognition datasets. The numbers in the table are computed using each similarity metric between the original and reconstructed images. Similar to the accuracy metric benchmark in \cref{tab:accuracy_1}, we evaluate the proposed method at different percentages of images sharing the same key model (1\%, 3\%, 10\%, and 70\%).

In the AgeDB-30 and LFW datasets, all similarity metrics improve as more images share a common key model. Interestingly, if only 3\% of the images share the same privacy key model, the quality of reconstructed images is comparable to that of the true key model. For example, in the AgeDB-30 dataset, the PSNR for 3\% shared key model images is 11.847 compared to 17.525 for the true key model images, and the MSE for 3\% shared key model images is 0.071 compared to 0.022 for the true key model images.

Furthermore, although pixel-based metrics such as PSNR and MSE show some differences between different percentages of images sharing the same key model, other metrics like LPIPS and CLIP do not show significant differences. For instance, in the AgeDB-30 dataset, the LPIPS for 3\% shared key model images is 0.450 compared to 0.367 for the true key model images, and the CLIP for 3\% shared key model images is 0.746 compared to 0.837 for the true key model images.

For the CFP-FP dataset, the reconstruction quality is relatively lower compared to the other two datasets. This is because the reconstruction quality with the true key model serves as a performance cap for the replicated key model, where the true key model's reconstruction quality is relatively lower. However, similar to the other two datasets, perceptual quality metrics such as LPIPS and CLIP still perform well. For example, in the CFP-FP dataset, the LPIPS for the true key model is 0.435, which is comparable to 0.623 for 1\% shared key model images. The CLIP metric also shows a consistent trend, with 0.825 for the true key model and 0.728 for 1\% shared key model images.

\begin{figure} 
    \centering
    \includegraphics[width=\if 1\doublecolumn 0.99 \else 0.5 \fi\linewidth]{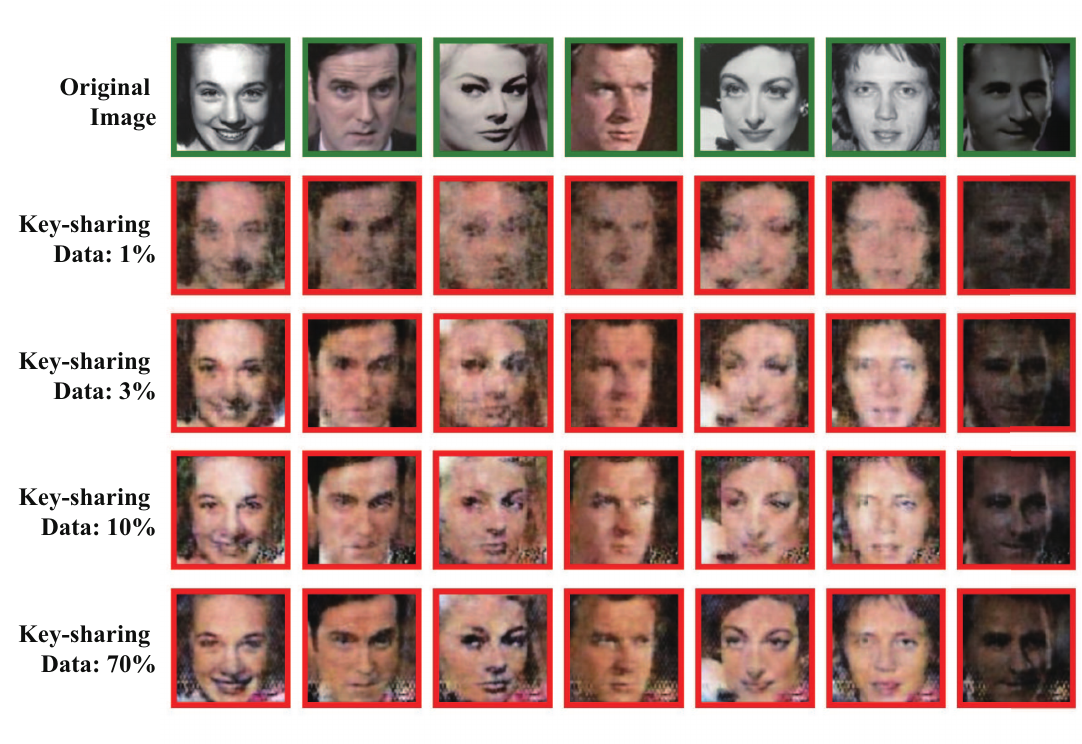}
    \caption{Graphical examples of the proposed method. The images in the first row with green borderlines show the original image and the images in the other rows show the reconstructed images by the proposed method. }
    \label{fig:graphical_example_main_1}
\end{figure}

\paragraph*{Graphical results}

In \cref{fig:graphical_example_main_1}, we present graphical examples of the proposed method. The first row with green borders shows the original images, while the subsequent rows display the reconstructed images with different percentages of leaked encrypted data (1\%, 3\%, 10\%, and 70\%).

With 1\% leaked encrypted data, the reconstructed images are significantly distorted and blurred, making recognition difficult, which aligns with lower similarity scores. As the percentage increases to 3\%, facial features become more distinguishable despite some blurring, showing noticeable improvement. At 10\%, the images are clearer and more recognizable, and at 70\%, the reconstructed images are very close to the original quality, supporting the highest similarity scores and demonstrating the method's effectiveness.

These results highlight that while pixel-based metrics like PSNR and MSE show improvement, perceptual similarity metrics such as LPIPS and CLIP also indicate significant enhancements in image quality (in \cref{tab:image_sim_1}). This improvement in reconstruction quality directly correlates with an increase in TPR (in \cref{tab:accuracy_1}), further validating the robustness of the proposed method in maintaining high image similarity and effective data reconstruction as more images share the same key model.

\subsection{Results for Various Face Recognition Models}

In \cref{tab:image_backbone_model}, we present the accuracy and similarity metrics for various backbones and face recognition schemes. Unlike the benchmarks in \cref{tab:accuracy_1,tab:image_sim_1}, the AdaFace and ArcFace face recognition models are used for evaluation, with their backbone configured as the IR-50 network.
As shown in the table, similar to the previous results, all evaluation metrics of the proposed method improve as more encrypted images share the same key model. For example, if 70\% of the gallery images share the same key model, the proposed method nearly achieves the reconstruction quality of the original key model. Interestingly, the proposed scheme can nearly achieve the perceptual similarity score of the original key model even when only 3\% of the gallery images share the key model.

This experiment demonstrates that the proposed method can be generally applied to various face recognition models and backbone networks.

\subsection{Ablation Study}

In this subsection, we aim to study the effect of our key contributions: 1) augmented identity loss and 2) GAN-based key model training. To this end, we implement the proposed method for all cases, whether the key contributions exist or not. In \cref{tab:main_face_table_2}, we present the quantitative results of our ablation study.
At lower percentages of leaked encrypted data, particularly 1\% and 3\%, we observe significant performance improvements due to the ablation study configurations. For example, incorporating both GAN loss and augmented identity loss at 1\% leaked data increases the TPR from 0.329 to 0.628 and reduces the MSE from 0.418 to 0.114. Similarly, for 3\% leaked data, the TPR increases from 0.299 to 0.746 and the MSE decreases from 0.437 to 0.071. These enhancements demonstrate the effectiveness of the ablation configurations in improving reconstruction quality when the amount of leaked data is minimal, aligning with the common belief regarding overfitting: less data leads to higher overfitting.

When the percentage of leaked encrypted data is higher, such as 10\% and 70\%, the performance improvements from ablation studies are relatively smaller but still notable. GAN-based training continues to enhance the performance metrics. For instance, at 70\% leaked data, the TPR increases from 0.860 to 0.929, and the MSE decreases from 0.193 to 0.063. This improvement is attributed to the increased amount of available data, which helps the reconstruction quality approach that of the original key model, thereby mitigating overfitting issues. Consequently, as more data becomes available, the model benefits from better generalization, leading to enhanced reconstruction fidelity.

\section{Discussion}
\label{sec:discussion}
\paragraph*{Conclusion}
This study investigates the potential vulnerabilities of the AVIH method~\cite{su2023hiding} by proposing a data reconstruction attack, highlighting the need for additional privacy protection methods in online image recognition systems. Our findings emphasize that if 1\% of the gallery dataset shares the same key model, the key model's functionality can be reconstructed, leading to a successful data reconstruction attack.

\paragraph*{Limitations and extensibility}

Although we implement our method for the AVIH method~\cite{su2023hiding}, it can be extended to other cloud-based machine learning systems where neural network-based key models are used for reconstructing original data. Since the work in \cite{su2023hiding} is recently published, few follow-up papers have appeared. However, we believe our method can be extended to all future works related to ML-based cloud-based systems.

\paragraph*{Future research direction}

Here, we discuss about defense method against our work.
One might consider assigning a unique neural network key model to each gallery image; however, this is extremely memory inefficient. 
Instead, we could assign an additional key image to each gallery image, where the original image can be reconstructed only when the key image and key model match exactly. 
If these images do not match, the reconstructed image would be another natural image.
None of the previous studies have proposed a defense method like this; however, since this is beyond the scope of our work, we leave this for future research.

\appendices

\section{Additional Experimental Results on Vehicle and Person Re-Identification}
\label{sec:additional_experiments}
\subsection{Implementation Details }

\paragraph*{Person \& vehicle re-identification scenario}

For image recognition tasks other than face images, we target to reconstruct the original images for the gallery set of the following datasets: vehicle re-identification dataset (VeRi~\cite{VeRIdataset}) and pedestrian re-identification dataset (Market-1501~\cite{zheng2015scalable}). 
For both dataset, the service model is chosen as TransReID model~\cite{he2021transreid} with ViT backbone~\cite{dosovitskiy2021an}. 
For the auxiliary datasets, we use Stanford Car~\cite{6755945} dataset and LPW~\cite{song_2018_rqen} dataset for vehicle/human images, respectively.
The accuracy on the re-identification is measured by mAP, rank-1, and rank-5 accuracy, where these metrics are evaluated by TransReID model with the DeiT backbone \cite{pmlr-v139-touvron21a}. 

\paragraph*{Implementation of AVIH and our data reconstruction attack}

The implementation details of the AVIH and the proposed method for re-identification tasks are similar to those in the face recognition experiment.  
The difference is that: the VC loss kernel size is configured as 8 for vehicle dataset, and our key model is trained for 800 steps.

\subsection{Accuracy Metrics}

\begin{table}[!htb]
    \setlength\tabcolsep{4pt} %
    \centering
    \caption{
    The accuracy measurements of the original gallery images, protected images, reconstructed images by original key model, and our data reconstruction attack results, where the datasets are VeRi (vehicle) and market-1501 (person) datasets.
    }
    \label{tab:reid_acc_1}
    \adjustbox{width=\if 1\doublecolumn .7 \else 0.45 \fi\linewidth}{
    \begin{tabular}{c|c|ccc}
    \toprule
         & & Rank-1 $\uparrow$ & Rank-5 $\uparrow$ & mAP $\uparrow$ \\
    \midrule
        \parbox[t]{2mm}{\multirow{7}{*}{\rotatebox[origin=c]{90}{Market-1501 (Person)}}} & Ours (1\%) & 0.015\textsuperscript{$\pm$0.015} & 0.045\textsuperscript{$\pm$0.040} & 0.025\textsuperscript{$\pm$0.019}  \\
         & Ours (3\%) & 0.147\textsuperscript{$\pm$0.033} & 0.296\textsuperscript{$\pm$0.061} & 0.151\textsuperscript{$\pm$0.033} \\
         & Ours (10\%) & 0.429\textsuperscript{$\pm$0.147} & 0.637\textsuperscript{$\pm$0.212} & 0.389\textsuperscript{$\pm$0.130} \\
         & Ours (70\%) & \textbf{0.772\textsuperscript{$\pm$0.013}} & \textbf{0.905\textsuperscript{$\pm$0.012}} & \textbf{0.685\textsuperscript{$\pm$0.015}} \\
    \cmidrule{2-5}
        & Original  & 0.912\textsuperscript{$\pm$0.000} & 0.969\textsuperscript{$\pm$0.000} & 0.830\textsuperscript{$\pm$0.000}\\
        & Protected  & 0.002\textsuperscript{$\pm$0.000} & 0.039\textsuperscript{$\pm$0.000} & 0.020\textsuperscript{$\pm$0.000} \\
        & Key model  & 0.881\textsuperscript{$\pm$0.000} & 0.953\textsuperscript{$\pm$0.000} & 0.797\textsuperscript{$\pm$0.000} \\
        \midrule
        \parbox[t]{2mm}{\multirow{7}{*}{\rotatebox[origin=c]{90}{VeRi (Vehicle)}}} & Ours (1\%) & 0.020\textsuperscript{$\pm$0.009} & 0.045\textsuperscript{$\pm$0.015} & 0.031\textsuperscript{$\pm$0.008} \\
         & Ours (3\%) & 0.052\textsuperscript{$\pm$0.030} & 0.109\textsuperscript{$\pm$0.059} & 0.060\textsuperscript{$\pm$0.028} \\
         & Ours (10\%) & 0.195\textsuperscript{$\pm$0.137} & 0.327\textsuperscript{$\pm$0.217} & 0.163\textsuperscript{$\pm$0.108} \\
         & Ours (70\%) & \textbf{0.520\textsuperscript{$\pm$0.064}} & \textbf{0.740\textsuperscript{$\pm$0.055}} & \textbf{0.406\textsuperscript{$\pm$0.044}} \\
    \cmidrule{2-5}
        & Original  & 0.900\textsuperscript{$\pm$0.000} & 0.976\textsuperscript{$\pm$0.000} & 0.719\textsuperscript{$\pm$0.000} \\
        & Protected  & 0.018\textsuperscript{$\pm$0.000} & 0.074\textsuperscript{$\pm$0.000} & 0.032\textsuperscript{$\pm$0.000} \\
        & Key model  & 0.707\textsuperscript{$\pm$0.000} & 0.874\textsuperscript{$\pm$0.000} & 0.529\textsuperscript{$\pm$0.000} \\
    \bottomrule
    \end{tabular}
    }
\end{table}

Table \ref{tab:reid_acc_1} presents the accuracy measurements for the original gallery images, protected images, reconstructed images by the original key model, and our data reconstruction attack results using the TransReID method with ViT backbone. The evaluations are conducted on two datasets: Market-1501 (Person) and VeRi (Vehicle).

For both datasets, all the accuracy matrics are enhanced as more gallery images share a common key model. 
For instance, in market-1501 results, if only 1\% of images share the same key model, the performance is quite poor, with a Rank-1 accuracy of 0.015, Rank-5 accuracy of 0.045, and mAP of 0.025. As the percentage of shared key models increases to 3\%, the performance improves significantly, with Rank-1 increasing to 0.147, Rank-5 to 0.296, and mAP to 0.151. When 70\% of the images share the same key model, the reconstructed images achieve near-original performance with Rank-1 at 0.772, Rank-5 at 0.905, and mAP at 0.685. 

Compared to the results in face recognition, the accuracy results are not good enough; however, as will be discussed later in the next subsection, the perceptual similarity of our proposed method closely achieves the images reconstructed by the private key model.

\begin{table}[tb]
    \setlength\tabcolsep{4pt} %
    \centering
    \caption{
    The reconstruction quality measurements of the original gallery images, protected images, reconstructed images by original key model, and our data reconstruction attack results.  
    }
    \label{tab:similarity_reid_1}
    \adjustbox{width=\if 1\doublecolumn .99 \else 0.7 \fi\linewidth}{
    \begin{tabular}{c|c|ccccc}
    \toprule
         & & MSE $\downarrow$ & LPIPS $\downarrow$ & PSNR $\uparrow$ & CLIP $\uparrow$ & SSIM $\uparrow$ \\
    \midrule
        \parbox[t]{2mm}{\multirow{7}{*}{\rotatebox[origin=c]{90}{Market-1501}}} & Ours (1\%) & 0.221\textsuperscript{$\pm$0.166} & 0.709\textsuperscript{$\pm$0.055} & 8.031\textsuperscript{$\pm$3.371} & 0.828\textsuperscript{$\pm$0.012} & 0.216\textsuperscript{$\pm$0.055} \\
         & Ours (3\%) & 0.078\textsuperscript{$\pm$0.009} & 0.557\textsuperscript{$\pm$0.023} & 11.170\textsuperscript{$\pm$0.507} & 0.853\textsuperscript{$\pm$0.009} & 0.296\textsuperscript{$\pm$0.013} \\
         & Ours (10\%) & 0.082\textsuperscript{$\pm$0.123} & 0.471\textsuperscript{$\pm$0.088} & 12.984\textsuperscript{$\pm$3.138} & 0.876\textsuperscript{$\pm$0.012} & 0.315\textsuperscript{$\pm$0.056} \\
         & Ours (70\%) & \textbf{0.033\textsuperscript{$\pm$0.005}} & \textbf{0.399\textsuperscript{$\pm$0.031}} & \textbf{14.986\textsuperscript{$\pm$0.649}} & \textbf{0.891\textsuperscript{$\pm$0.007}} & \textbf{0.351\textsuperscript{$\pm$0.019}} \\
    \cmidrule{2-7}
        & Original  & 0.000\textsuperscript{$\pm$0.000} & 0.009\textsuperscript{$\pm$0.000} & 36.831\textsuperscript{$\pm$0.000} & 0.980\textsuperscript{$\pm$0.000} & 0.492\textsuperscript{$\pm$0.000} \\
        & Protected  & 0.409\textsuperscript{$\pm$0.000} & 1.438\textsuperscript{$\pm$0.000} & 3.945\textsuperscript{$\pm$0.000} & 0.571\textsuperscript{$\pm$0.000} & 0.015\textsuperscript{$\pm$0.000} \\
        & Key model  & 0.003\textsuperscript{$\pm$0.000} & 0.408\textsuperscript{$\pm$0.000} & 25.447\textsuperscript{$\pm$0.000} & 0.927\textsuperscript{$\pm$0.000} & 0.422\textsuperscript{$\pm$0.000} \\
        \midrule
        \parbox[t]{2mm}{\multirow{7}{*}{\rotatebox[origin=c]{90}{VeRi}}} & Ours (1\%) & 0.288\textsuperscript{$\pm$0.184} & 0.787\textsuperscript{$\pm$0.026} & 6.434\textsuperscript{$\pm$2.743} & 0.715\textsuperscript{$\pm$0.003} & 0.167\textsuperscript{$\pm$0.028} \\
         & Ours (3\%) & 0.228\textsuperscript{$\pm$0.210} & 0.720\textsuperscript{$\pm$0.048} & 7.864\textsuperscript{$\pm$3.032} & 0.731\textsuperscript{$\pm$0.007} & 0.201\textsuperscript{$\pm$0.034} \\
         & Ours (10\%) & 0.297\textsuperscript{$\pm$0.253} & 0.668\textsuperscript{$\pm$0.080} & 7.209\textsuperscript{$\pm$3.850} & 0.751\textsuperscript{$\pm$0.027} & 0.215\textsuperscript{$\pm$0.059} \\
         & Ours (70\%) & \textbf{0.083\textsuperscript{$\pm$0.022}} & \textbf{0.598\textsuperscript{$\pm$0.025}} & \textbf{11.059\textsuperscript{$\pm$1.079}} & \textbf{0.793\textsuperscript{$\pm$0.009}} & \textbf{0.267\textsuperscript{$\pm$0.014}} \\
    \cmidrule{2-7}
        & Original  & 0.000\textsuperscript{$\pm$0.000} & 0.010\textsuperscript{$\pm$0.000} & 34.907\textsuperscript{$\pm$0.000} & 0.975\textsuperscript{$\pm$0.000} & 0.487\textsuperscript{$\pm$0.000} \\
        & Protected  & 0.410\textsuperscript{$\pm$0.000} & 1.385\textsuperscript{$\pm$0.000} & 3.963\textsuperscript{$\pm$0.000} & 0.593\textsuperscript{$\pm$0.000} & 0.017\textsuperscript{$\pm$0.000} \\
        & Key model  & 0.026\textsuperscript{$\pm$0.000} & 0.759\textsuperscript{$\pm$0.000} & 16.172\textsuperscript{$\pm$0.000} & 0.867\textsuperscript{$\pm$0.000} & 0.276\textsuperscript{$\pm$0.000} \\
    \bottomrule
    \end{tabular}
    }
\end{table}

\subsection{Similarity Metrics}

In \cref{tab:similarity_reid_1}, the evaluation of the reconstructed images on both datasets is presented using perceptual similarity and pixel-based similarity metrics. For the Market-1501 dataset, the MSE decreases as more images share a common key model, with the lowest MSE of 0.033 for 70\% shared key model images. Although pixel-based metrics like PSNR improve significantly from 8.031 (1\%) to 14.986 (70\%), perceptual similarity metrics such as LPIPS and CLIP also show significant enhancement, with LPIPS decreasing from 0.709 to 0.399 and CLIP increasing from 0.828 to 0.891.

Similarly, in the VeRi dataset, both perceptual and pixel-based similarity metrics show improvement. For instance, the PSNR increases from 6.434 (1\%) to 11.059 (70\%), while LPIPS decreases from 0.787 to 0.598 and CLIP increases from 0.715 to 0.793. These improvements in perceptual metrics indicate that the reconstructed images, even with a higher percentage of shared key models, maintain a high level of visual similarity to the original images.

These results suggest that while pixel-based metrics like PSNR and MSE improve, perceptual similarity metrics such as LPIPS and CLIP also indicate significant enhancements in image quality even nearly achieves the original key model. 
This shows that high image similarity and effective privacy protection as more images share the same key model.

\begin{figure}[!t]
    \centering
    \subfloat[Market-1501 (Person) Dataset]{\includegraphics[width=\if 1\doublecolumn .95 \else 0.6 \fi\linewidth]{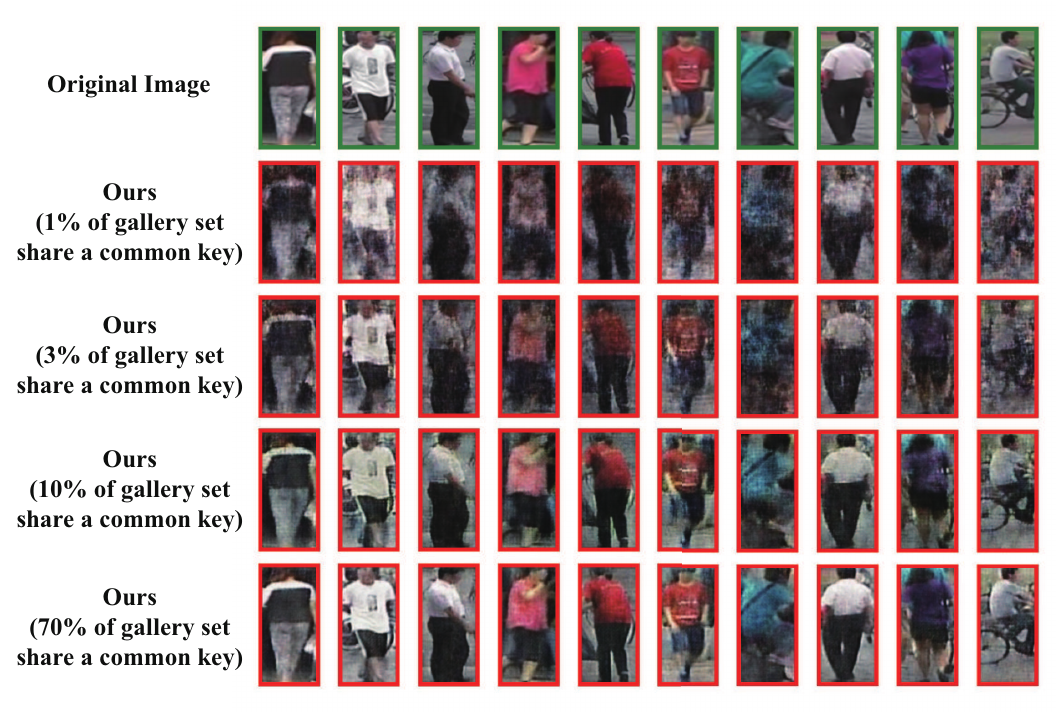}\label{subfig:graphical_example_main_2_1}} \\ 
    \subfloat[VeRi (Vehicle) Dataset]{\includegraphics[width=\if 1\doublecolumn .95 \else 0.6 \fi\linewidth]{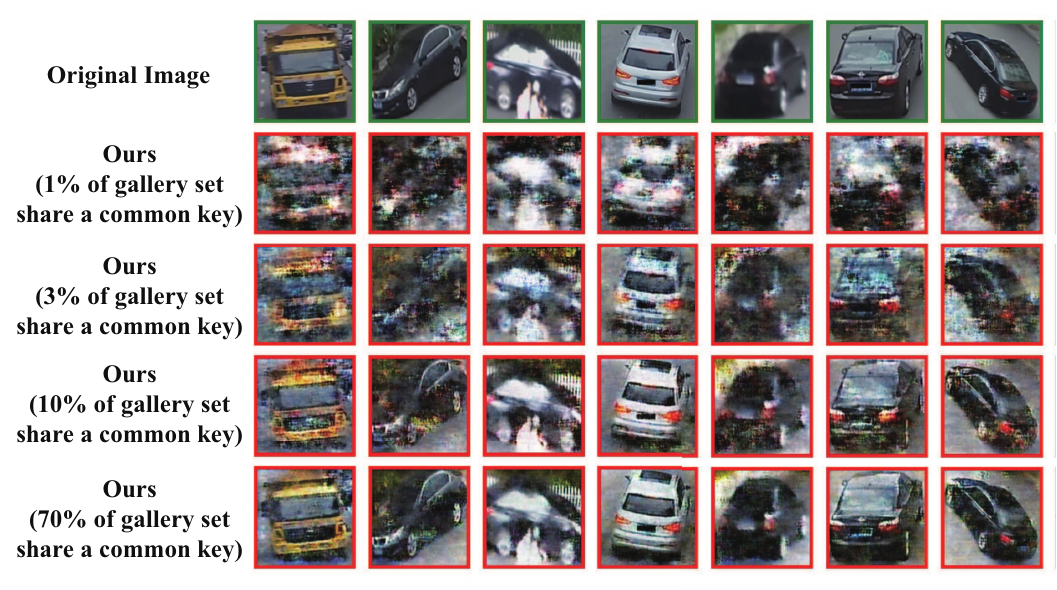}\label{subfig:graphical_example_main_2_2}}
    \caption{Examples of the reconstructed images by the proposed method. The index of each image is randomly chosen. The images with the green-rectangular borderline are the original images. In the re-identification task, we evaluate two datasets: (a) Market-1501 dataset and (b) VeRi Dataset.}
    \label{fig:graphical_example_main_2}
\end{figure}

\subsection{Graphical Results}

In \cref{fig:graphical_example_main_2}, we present graphical examples of the proposed method. The first row with green borderlines shows the original images, while the subsequent rows display the reconstructed images with different percentages of leaked encrypted data (1\%, 3\%, 10\%, and 70\%).

With 1\% leaked encrypted data, the reconstructed images are significantly distorted and blurred, making recognition difficult, which aligns with lower similarity scores. As the percentage increases to 3\%, facial features become more distinguishable despite some blurring, showing noticeable improvement. At 10\%, the images are clearer and more recognizable, and at 70\%, the reconstructed images are very close to the original quality, supporting the highest similarity scores and demonstrating the method's effectiveness.

\bibliographystyle{IEEEtran}
\bibliography{egbib}

\begin{thebibliography}{10}
\providecommand{\url}[1]{#1}
\csname url@samestyle\endcsname
\providecommand{\newblock}{\relax}
\providecommand{\bibinfo}[2]{#2}
\providecommand{\BIBentrySTDinterwordspacing}{\spaceskip=0pt\relax}
\providecommand{\BIBentryALTinterwordstretchfactor}{4}
\providecommand{\BIBentryALTinterwordspacing}{\spaceskip=\fontdimen2\font plus
\BIBentryALTinterwordstretchfactor\fontdimen3\font minus
  \fontdimen4\font\relax}
\providecommand{\BIBforeignlanguage}[2]{{%
\expandafter\ifx\csname l@#1\endcsname\relax
\typeout{** WARNING: IEEEtran.bst: No hyphenation pattern has been}%
\typeout{** loaded for the language `#1'. Using the pattern for}%
\typeout{** the default language instead.}%
\else
\language=\csname l@#1\endcsname
\fi
#2}}
\providecommand{\BIBdecl}{\relax}
\BIBdecl

\bibitem{su2023hiding}
Z.~Su, D.~Zhou, N.~Wang, D.~Liu, Z.~Wang, and X.~Gao, ``Hiding visual
  information via obfuscating adversarial perturbations,'' in \emph{Int. Conf.
  Comput. Vis. (ICCV)}, 2023, pp. 4356--4366.

\bibitem{chollet2024privacypreservingpersonalassistant}
G.~Chollet, H.~Sansen, Y.~Tevissen, J.~Boudy, M.~Hariz, C.~Lohr, and F.~Yassa,
  ``Privacy preserving personal assistant with on-device diarization and spoken
  dialogue system for home and beyond,'' \emph{arXiv preprint
  arXiv:2401.01146}, 2024.

\bibitem{singh2017cloud}
A.~Singh and K.~Chatterjee, ``Cloud security issues and challenges: A survey,''
  \emph{J. Netw. Comput. Appl.}, vol.~79, pp. 88--115, 2017.

\bibitem{he2020fastreid}
L.~He, X.~Liao, W.~Liu, X.~Liu, P.~Cheng, and T.~Mei, ``Fast{ReID}: A pytorch
  toolbox for general instance re-identification,'' \emph{arXiv preprint
  arXiv:2006.02631}, 2020.

\bibitem{he2021transreid}
S.~He, H.~Luo, P.~Wang, F.~Wang, H.~Li, and W.~Jiang, ``Trans{ReID}:
  Transformer-based object re-identification,'' in \emph{Int. Conf. Comput.
  Vis. (ICCV)}, 2021, pp. 15\,013--15\,022.

\bibitem{zhu2018hidden}
J.~Zhu, R.~Kaplan, J.~Johnson, and L.~Fei-Fei, ``Hidden: Hiding data with deep
  networks,'' in \emph{Eur. Conf. Comput. Vis. (ECCV)}, 2018, pp. 657--672.

\bibitem{xiang2019visual}
T.~Xiang, Y.~Yang, H.~Liu, and S.~Guo, ``Visual security evaluation of
  perceptually encrypted images based on image importance,'' \emph{IEEE Trans.
  Circuit Syst. Video Technol.}, vol.~30, no.~11, pp. 4129--4142, 2019.

\bibitem{guo2019peid}
S.~Guo, T.~Xiang, X.~Li, and Y.~Yang, ``{PEID}: A perceptually encrypted image
  database for visual security evaluation,'' \emph{IEEE Trans. Inf. Forensics
  Secur.}, vol.~15, pp. 1151--1163, 2019.

\bibitem{lee2022low}
E.~Lee, J.-W. Lee, J.~Lee, Y.-S. Kim, Y.~Kim, J.-S. No, and W.~Choi,
  ``Low-complexity deep convolutional neural networks on fully homomorphic
  encryption using multiplexed parallel convolutions,'' in \emph{Int. Conf.
  Mach. Learn. (ICML)}.\hskip 1em plus 0.5em minus 0.4em\relax PMLR, 2022, pp.
  12\,403--12\,422.

\bibitem{kim2023optimized}
D.~Kim and C.~Guyot, ``Optimized privacy-preserving cnn inference with fully
  homomorphic encryption,'' \emph{IEEE Trans. Inf. Forensics Secur.}, vol.~18,
  pp. 2175--2187, 2023.

\bibitem{Nguyen_2023_CVPR}
N.-B. Nguyen, K.~Chandrasegaran, M.~Abdollahzadeh, and N.-M. Cheung,
  ``Re-thinking model inversion attacks against deep neural networks,'' in
  \emph{IEEE Conf. Comput. Vis. Pattern Recog. (CVPR)}, June 2023.

\bibitem{Tian_23_roleof}
Z.~Tian, L.~Cui, C.~Zhang, S.~Tan, S.~Yu, and Y.~Tian, ``The role of class
  information in model inversion attacks against image deep learning
  classifiers,'' \emph{IEEE Trans. Dependable Secure Comput.}, pp. 1--14, 2023.

\bibitem{Zhou_23_boosting_MI_adv}
S.~Zhou, T.~Zhu, D.~Ye, X.~Yu, and W.~Zhou, ``Boosting model inversion attacks
  with adversarial examples,'' \emph{IEEE Trans. Dependable Secure Comput.},
  pp. 1--18, 2023.

\bibitem{hu2022membership}
H.~Hu, Z.~Salcic, L.~Sun, G.~Dobbie, P.~S. Yu, and X.~Zhang, ``Membership
  inference attacks on machine learning: A survey,'' \emph{ACM Computing
  Surveys (CSUR)}, vol.~54, no. 11s, pp. 1--37, 2022.

\bibitem{Fredrikson14}
M.~Fredrikson, E.~Lantz, S.~Jha, S.~Lin, D.~Page, and T.~Ristenpart, ``Privacy
  in pharmacogenetics: An end-to-end case study of personalized warfarin
  dosing,'' in \emph{Proceedings of USENIX Conference on Security Symposium},
  2014, pp. 17--32.

\bibitem{Fredrikson15}
M.~Fredrikson, S.~Jha, and T.~Ristenpart, ``Model inversion attacks that
  exploit confidence information and basic countermeasures,'' in
  \emph{Proceedings of ACM SIGSAC Conf. Computer and Commun. Secur.}, 2015, pp.
  1322--1333.

\bibitem{Hidano17}
S.~Hidano, T.~Murakami, S.~Katsumata, S.~Kiyomoto, and G.~Hanaoka, ``Model
  inversion attacks for prediction systems: Without knowledge of non-sensitive
  attributes,'' in \emph{Proceedings of Annual Conference on Privacy, Security
  and Trust}, 2017, pp. 115--11\,509.

\bibitem{jang2023patchmi}
J.~Jang, H.~Lyu, and H.~J. Yang, ``Patch-{MI}: Enhancing model inversion
  attacks via patch-based reconstruction,'' \emph{arXiv preprint
  arxiv:2312.07040}, 2023.

\bibitem{haim2022reconstructing}
N.~Haim, G.~Vardi, G.~Yehudai, O.~Shamir, and M.~Irani, ``Reconstructing
  training data from trained neural networks,'' \emph{Adv. Neural Inf. Process.
  Syst. (NeurIPS)}, vol.~35, pp. 22\,911--22\,924, 2022.

\bibitem{gilad2016cryptonets}
R.~Gilad-Bachrach, N.~Dowlin, K.~Laine, K.~Lauter, M.~Naehrig, and J.~Wernsing,
  ``Crypto{N}ets: Applying neural networks to encrypted data with high
  throughput and accuracy,'' in \emph{Int. Conf. Mach. Learn. (ICML)}.\hskip
  1em plus 0.5em minus 0.4em\relax PMLR, 2016, pp. 201--210.

\bibitem{lee2022privacy}
J.-W. Lee, H.~Kang, Y.~Lee, W.~Choi, J.~Eom, M.~Deryabin, E.~Lee, J.~Lee,
  D.~Yoo, Y.-S. Kim \emph{et~al.}, ``Privacy-preserving machine learning with
  fully homomorphic encryption for deep neural network,'' \emph{IEEE Access},
  vol.~10, pp. 30\,039--30\,054, 2022.

\bibitem{ding2020deepedn}
Y.~Ding, G.~Wu, D.~Chen, N.~Zhang, L.~Gong, M.~Cao, and Z.~Qin, ``Deepedn: A
  deep-learning-based image encryption and decryption network for internet of
  medical things,'' \emph{IEEE Internet Things J.}, vol.~8, no.~3, pp.
  1504--1518, 2020.

\bibitem{sirichotedumrong2021gan}
W.~Sirichotedumrong and H.~Kiya, ``A gan-based image transformation scheme for
  privacy-preserving deep neural networks,'' in \emph{European Signal Process.
  Conf. (EUSIPCO)}.\hskip 1em plus 0.5em minus 0.4em\relax IEEE, 2021, pp.
  745--749.

\bibitem{chuman2023jigsaw}
T.~Chuman and H.~Kiya, ``A jigsaw puzzle solver-based attack on block-wise
  image encryption for privacy-preserving dnns,'' in \emph{Int. Workshop on
  Adv. Imaging Technol. (IWAIT)}, vol. 12592.\hskip 1em plus 0.5em minus
  0.4em\relax SPIE, 2023, pp. 335--340.

\bibitem{maungmaung2023generative}
A.~MaungMaung and H.~Kiya, ``Generative model-based attack on learnable image
  encryption for privacy-preserving deep learning,'' \emph{arXiv preprint
  arXiv:2303.05036}, 2023.

\bibitem{tang2023watermarking}
Y.~Tang, J.~Yu, K.~Gai, X.~Qu, Y.~Hu, G.~Xiong, and Q.~Wu, ``Watermarking
  vision-language pre-trained models for multi-modal embedding as a service,''
  \emph{arxiv preprint arxiv:2311.05863}, 2023.

\bibitem{qiao2023novel}
T.~Qiao, Y.~Ma, N.~Zheng, H.~Wu, Y.~Chen, M.~Xu, and X.~Luo, ``A novel model
  watermarking for protecting generative adversarial network,'' \emph{Computers
  \& Security}, vol. 127, p. 103102, 2023.

\bibitem{baluja2017hiding}
S.~Baluja, ``Hiding images in plain sight: Deep steganography,'' \emph{Adv.
  Neural Inf. Process. Syst. (NeurIPS)}, vol.~30, 2017.

\bibitem{deng2019arcface}
J.~Deng, J.~Guo, N.~Xue, and S.~Zafeiriou, ``Arcface: Additive angular margin
  loss for deep face recognition,'' in \emph{IEEE Conf. Comput. Vis. Pattern
  Recog. (CVPR)}, 2019, pp. 4690--4699.

\bibitem{kim2022adaface}
M.~Kim, A.~K. Jain, and X.~Liu, ``Adaface: Quality adaptive margin for face
  recognition,'' in \emph{IEEE Conf. Comput. Vis. Pattern Recog. (CVPR)}, 2022,
  pp. 18\,750--18\,759.

\bibitem{goodfellow2014generative}
I.~Goodfellow, J.~Pouget-Abadie, M.~Mirza, B.~Xu, D.~Warde-Farley, S.~Ozair,
  A.~Courville, and Y.~Bengio, ``Generative adversarial nets,'' \emph{Adv.
  Neural Inf. Process. Syst. (NeurIPS)}, vol.~27, 2014.

\bibitem{LFWTech}
G.~B. Huang, M.~Ramesh, T.~Berg, and E.~Learned-Miller, ``Labeled faces in the
  wild: A database for studying face recognition in unconstrained
  environments,'' University of Massachusetts, Amherst, Tech. Rep. 07-49,
  October 2007.

\bibitem{moschoglou2017agedb}
S.~Moschoglou, A.~Papaioannou, C.~Sagonas, J.~Deng, I.~Kotsia, and
  S.~Zafeiriou, ``Age{DB}: The first manually collected, in-the-wild age
  database,'' in \emph{IEEE Conf. Comput. Vis. Pattern Recog. Worksh.}, 2017.

\bibitem{cfp-paper}
S.~Sengupta, J.-C. Chen, C.~Castillo, V.~M. Patel, R.~Chellappa, and D.~W.
  Jacobs, ``Frontal to profile face verification in the wild,'' in \emph{Winter
  Conf. Appl. Comput. Vis. (WACV)}, 2016, pp. 1--9.

\bibitem{liu2015faceattributes}
Z.~Liu, P.~Luo, X.~Wang, and X.~Tang, ``Deep learning face attributes in the
  wild,'' in \emph{Int. Conf. Comput. Vis. (ICCV)}, December 2015.

\bibitem{Unet}
O.~Ronneberger, P.~Fischer, and T.~Brox, ``{U-Net}: Convolutional networks for
  biomedical image segmentation,'' \emph{CoRR}, vol. abs/1505.04597, 2015.

\bibitem{zhang2018perceptual}
R.~Zhang, P.~Isola, A.~A. Efros, E.~Shechtman, and O.~Wang, ``The unreasonable
  effectiveness of deep features as a perceptual metric,'' in \emph{IEEE Conf.
  Comput. Vis. Pattern Recog. (CVPR)}, 2018.

\bibitem{hessel2021clipscore}
J.~Hessel, A.~Holtzman, M.~Forbes, R.~L. Bras, and Y.~Choi, ``{CLIPScore:} a
  reference-free evaluation metric for image captioning,'' in \emph{Empirical
  Methods Nat. Lang. Process.}, 2021.

\bibitem{1284395}
Z.~Wang, A.~Bovik, H.~Sheikh, and E.~Simoncelli, ``Image quality assessment:
  from error visibility to structural similarity,'' \emph{IEEE Trans. Image
  Process.}, vol.~13, no.~4, pp. 600--612, 2004.

\bibitem{VeRIdataset}
X.~Liu, W.~Liu, T.~Mei, and H.~Ma, ``{PROVID}: Progressive and multimodal
  vehicle reidentification for large-scale urban surveillance,'' \emph{IEEE
  Trans. Multimedia}, vol.~20, no.~3, pp. 645--658, 2018.

\bibitem{zheng2015scalable}
L.~Zheng, L.~Shen, L.~Tian, S.~Wang, J.~Wang, and Q.~Tian, ``Scalable person
  re-identification: A benchmark,'' in \emph{Int. Conf. Comput. Vis. (ICCV)},
  2015.

\bibitem{dosovitskiy2021an}
A.~Dosovitskiy \emph{et~al.}, ``An image is worth 16x16 words: Transformers for
  image recognition at scale,'' in \emph{Int. Conf. Learn. Represent. (ICLR)},
  2021.

\bibitem{6755945}
J.~Krause, M.~Stark, J.~Deng, and L.~Fei-Fei, ``{3D} object representations for
  fine-grained categorization,'' in \emph{Int. Conf. Comput. Vis. Worksh.},
  2013, pp. 554--561.

\bibitem{song_2018_rqen}
G.~Song, B.~Leng, Y.~Liu, C.~Hetang, and S.~Cai, ``Region-based quality
  estimation network for large-scale person re-identification,'' in
  \emph{Assoc. Adv. Artif. Intell. (AAAI)}, 2018.

\bibitem{pmlr-v139-touvron21a}
H.~Touvron, M.~Cord, M.~Douze, F.~Massa, A.~Sablayrolles, and H.~Jegou,
  ``Training data-efficient image transformers \& distillation through
  attention,'' in \emph{Int. Conf. Mach. Learn. (ICML)}, 2021.

\end{thebibliography}

\end{document}